\documentclass{sig-alternate-05-2015}
\makeatletter
\def\@copyrightspace{\relax}
\makeatother

\newcommand{\totalDatasets}{45}
\newcommand{\tunedWins}{35}
\newcommand{\fixedWins}{34}

\usepackage[hidelinks,pdfpagelabels=false]{hyperref}

\hypersetup{
    colorlinks = true,
	allcolors = blue
}

\usepackage{cite}

\usepackage{algorithm}

\usepackage{algpseudocode}

\usepackage{graphicx}
\graphicspath{{./figures/}}    
\DeclareGraphicsExtensions{.pdf,.jpeg,.png}

\usepackage[caption=false]{subfig}

\usepackage{multirow}
\usepackage[table]{xcolor}
\usepackage{adjustbox}
\usepackage{alphalph}

\begin{document}

\title{Ensembles of Randomized Time Series Shapelets Provide Improved Accuracy while Reducing Computational Costs}
\numberofauthors{2} 
\author{
\alignauthor Atif Raza\\
        \email{atifraza@uni-mainz.de}
\alignauthor Stefan Kramer\\
        \email{kramer@informatik.uni-mainz.de}
\end{tabular}\newline\newline\begin{tabular}{c}
	\affaddr{Institute of Computer Science}\\
	\affaddr{Johannes Gutenberg University Mainz}\\
	\affaddr{Mainz, Germany}
}

\maketitle
\begin{abstract}
Shapelets are discriminative time series subsequences that allow generation of interpretable classification models, which provide faster and generally better classification than the nearest neighbor approach.
However, the shapelet discovery process requires the evaluation of all possible subsequences of all time series in the training set, making it extremely computation intensive.
Consequently, shapelet discovery for large time series datasets quickly becomes intractable.
A number of improvements have been proposed to reduce the training time.
These techniques use approximation or discretization and often lead to reduced classification accuracy compared to the exact method.

We are proposing the use of ensembles of shapelet-based classifiers obtained using random sampling of the shapelet candidates.
Using random sampling reduces the number of evaluated candidates and consequently the required computational cost, while the classification accuracy of the resulting models is also not significantly different than that of the exact algorithm.
The combination of randomized classifiers rectifies the inaccuracies of individual models because of the diversity of the solutions.
Based on the experiments performed, it is shown that the proposed approach of using an ensemble of inexpensive classifiers provides better classification accuracy compared to the exact method at a significantly lesser computational cost.
\end{abstract}

\section{Introduction}\label{section:intro}
Time series data mining, in general, and classification, in particular, has seen a huge interest. 
The most investigated approach for time series classification has been the nearest neighbor algorithm coupled with various distance measures \cite{Ding2008CompDistMeasure, Bagnall2014NNEval}.
The nearest neighbor approach is simple to implement and requires little to no parameter tuning.\footnote{The distance measure used may require parameter tuning; e.g.\ DTW requires a window parameter for optimal results.}
However, it also has a few drawbacks.
It requires the storage of the entire training set with instances belonging to all the classes and the time required for classification is directly proportional to the number of instances in the training set.
It also fails to provide a clear insight about why a particular instance was classified as belonging to a certain class except that it was a ``best match'' to some instance of the assigned class.

Time series shapelets (\textit{YK-Shapelets}) were introduced to address the drawbacks of nearest neighbor based time series classification \cite{Ye2009Shapelets}.
In the nearest neighbor approach, full-length time series are compared and an instance is classified as belonging to the class of the best matching instance, or nearest neighbor, from the training set.
For time series shapelets, instead of comparing entire shapes, the presence of small subsequences unique to a particular class is sought.
Therefore, a shapelet is the most representative subsequence occurring in the instances of a certain class and its presence in a new instance leads to the classification of that instance as belonging to that particular class.
This is also the aspect which gives time series shapelets greater insight than the nearest neighbor approach because we can state that the time series in question was classified as such because of the presence (or absence) of a particular shapelet.
For example, we can differentiate between normal and abnormal patterns of an ECG time series based on the presence of certain shapelets and hence, classify the ECG data for a healthy or unhealthy person and also identify the particular heart disease.
The process of shapelet discovery can be divided into
(1) the enumeration of subsequences of all possible lengths for all the instances in the training set, and
(2) the evaluation of all the subsequences to find the one(s) capable of dividing the dataset as best as possible and preferably, provide subsets containing only the instances of a single class.

Shapelets based classification has  many advantages over the nearest neighbor approach.
It provides better insight into the classification process and the results can be verified by domain experts.
It can provide a better understanding of the data and may even discover unknown artifacts, providing new information and contributing fresh knowledge.
It is also much faster than the nearest neighbor approach because it only searches for a single subsequence in the incoming time series whereas a (full length) comparison with all training instances is required for the nearest neighbor approach.
Finally, shapelets are local features, so they can be significantly more accurate for certain problems because time series classification using global features can be highly susceptible to noise and distortions.

Despite its many advantages, the huge computational cost of shapelets based classification is a major drawback of this technique.
For a dataset with $k$ instances of length $m$, the number of all possible shapelet candidates of all lengths is $\frac{1}{2}km(m+1)$, which is on the order of $O(km^2)$.
Evaluating each candidate requires its comparison with $O(km)$ candidates on average and each comparison (using Euclidean distance) on average requires $O(m)$ operations.
Therefore, the brute force approach has a time complexity of $O(k^2m^4)$.
Clearly, the computational complexity of training a shapelets based classifier is untenable even for very small datasets.
However, a number of techniques have been proposed to reduce the time complexity of the shapelet discovery process.

Ensembles of machine learning models have been shown to often outperform individual models, provided the models constituting the ensemble are
(1) accurate, i.e.\ provide better results than random guessing, and
(2) diverse, i.e.\ make different errors for an unseen problem \cite{Hansen1990NNEnsembles}.
Ensembles enable the use of a number of ``inexpensive'' models instead of a single, expensive and highly accurate model.
Therefore, we are proposing the use of ensembles of inexpensive shapelet-based classifiers.
This approach can provide better classification accuracy as compared to the YK-Shapelets method at a reduced computational cost.
Varying the number of classifiers in the ensemble provides a mechanism to obtain highly accurate or computationally less expensive models.
The averaging behavior of ensembles also reduces the variance of the individual models.


\section{Background}\label{section:background}
The vast amounts of time series data are a treasure trove of information waiting to be mined and explored for the hidden insights they can provide.
Time series classification using the nearest neighbor approach is a very simple and highly effective technique that has been used extensively.
However, the computational complexity of the classification phase along with little to no interpretability has lead to a subsequence or shapelets based classification approach.
In shapelets based classification, the presence (or absence) of a specific subsequence, or ``shapelet'', in a time series determines its class.
If the distance of a time series from the shapelet is less than a threshold, then it is said to contain that shapelet and vice versa.
We will summarize the shapelet discovery algorithm and some of the proposed speed-up techniques in sections \ref{section:shapelet_discovery} and \ref{section:speed_up_techniques}.
For a detailed introduction, we refer the readers to the respective papers \cite{Ye2009Shapelets, Mueen2011Logical, Rakthanmanon2013FastShapelets, Renard2015RandomShapelets, Karlsson2016gRSF}.

A \textit{time series} $T=t_1,t_2,\ldots,t_m$, is an ordered set of $m$ real-valued features.
A \textit{subsequence} $S$ of length $l$ is a contiguous chunk of a time series, such that $S=t_p,t_{p+1},\ldots,t_{p+l-1}$, for $1 \leq p \leq m-l+1$.
For a time series $T$, the set of all subsequences of length $l$ is given as $S_T^l=\{S_1^l, S_2^l, \ldots, S_{m-l+1}^l\}$, where the subscript denotes the starting position of the subsequence.
The number of possible subsequences of length $l$ in a time series of length $m$ is equal to $m-l+1$.
The number of subsequences in a dataset consisting of $k$ instances of length $m$ and possible shapelet candidate lengths in the range $[min, max]$ is equal to:
\begin{equation*}
\sum_{l = min}^{max}\sum_{i = 1}^{k}(m-l+1)
\end{equation*}
Since the shapelet discovery process is used to split the dataset into purer splits and create a decision tree based on shapelets and their corresponding distance thresholds, the number of candidates stated above is only for the first call to the shapelet discovery process.
Subsequent calls further increase the number of evaluated candidates, although the number of candidates decreases at each node.

The distance between two subsequences of length $l$ is defined as $dist(S,R)=\sum_{i=1}^l(s_i-r_i)^2$, which is simply the Euclidean distance without the square root.
The distance between a subsequence $S$ and a time series $T$ is defined as the minimum distance between $S$ and all subsequences of $T$ having length $|S|$ i.e.\ $dist(S,T)=min(dist(S,S')), \forall S' \in S_T^{|S|}$.

\subsection{Shapelet Discovery}\label{section:shapelet_discovery}
The shapelet discovery algorithm aims to split the dataset into ``purer'' subsets such that the instances with and without the shapelet form two separate splits.
The purity of the obtained splits is evaluated using the information gain measure although other approaches can be used as well \cite{Lines2012ShapeletTransform}.
The shapelet discovery process is embedded in a decision tree learner.
At each node, the algorithm searches for the shapelet and split distance pair, which maximizes the information gain when used to split the dataset.
This shapelet and split distance pair constitutes the decision criterion for the particular node and splits the dataset for subsequent nodes of the tree.
The decision tree learner initiates the shapelet discovery process or forms leaf nodes with the splits depending on their purity levels.
Algorithm \ref{algo:decision-tree} provides the basic algorithm for learning the classification model.

\begin{algorithm}[tb]
    \caption{CreateTree ($D$, $l$, $u$)}
    \label{algo:decision-tree}
    \begin{algorithmic}[1]
        \Require $D$: Time Series dataset, $l$/$u$: min/max length of shapelet candidates
        \Ensure Shapelets based decision tree $DT$
        \If{\Call{IsPure}{D}}
        \State \textbf{return} \Call{CreateLeafNode}{D}
        \Else
        \State $(\mathcal{S},\delta,d_{map})\gets$ \Call{FindShapelet}{$D,l,u$}
        \State $(D_L,D_R)\gets$ \Call{SplitData}{$D,\delta,d_{map}$}
        \State ${node}_L \gets$ \Call{CreateTree}{$D_L,l,u$}
        \State ${node}_R \gets$ \Call{CreateTree}{$D_R,l,u$}
        \State \textbf{return} $(\mathcal{S}, \delta, {node}_L, {node}_R)$
        \EndIf
    \end{algorithmic}
\end{algorithm}

Algorithm \ref{algo:YKAlgo} lists the brute force shapelet discovery process.
It takes the time series dataset $D$ and the parameters $minLen$ and $maxLen$ as inputs and loops through the length parameters to evaluate all possible candidates (Line \ref{YKAlgo:FirstLoop}).
The \textit{GetNextCandidate} procedure (Line \ref{YKAlgo:GenerateCandidate}) provides the next shapelet candidate for evaluation or returns an empty candidate signaling candidate exhaustion for the current length.
It keeps track of the current instance number and the starting point for the shapelet candidate to sequentially generate new candidates using the current candidate length, removing the requirement of creating the candidates beforehand.
In Lines \ref{YKAlgo:GetDistancesStart} to \ref{YKAlgo:GetDistancesEnd}, the distance between the current candidate and each time series instance in the dataset is obtained and an order line is created.
The best information gain and the corresponding splitting distance are obtained for the order line (Line \ref{YKAlgo:CheckCandidate}).
If the new information gain is greater than the best so far, the values for best so far information gain, split distance, shapelet and order line are updated.
After evaluating all candidates, the best-found shapelet along with the corresponding split distance and order line are returned.
The shapelet and split distance constitute the decision criterion of the node.
The dataset is split using the split distance and order line for subsequent nodes.
When a split reaches the required purity level, a leaf node is created with the majority class of the instances reaching that node, otherwise the shapelet discovery process is called for the new split.

\begin{algorithm}[tb]
    \caption{FindShapelet ($D$, $minLen$, $maxLen$)}
    \label{algo:YKAlgo}
    \begin{algorithmic}[1]
        \Require $D$: Time Series Dataset, $minLen$: minimum candidate length, $maxLen$: maximum candidate length
        \Ensure $\mathcal{S}$: Shapelet, $\delta$: split distance, $d_{map}$: distance line
        \State $bsf\_InfoGain \gets -\infty$
        \State $bsf\_SplitDist \gets -\infty$
        \State $bsf\_OrderLine \gets \varnothing$
        \For{$len = maxLen$ \textbf{to} $minLen$}\label{YKAlgo:FirstLoop}
        \State $order\_line \gets \varnothing$
        \While{$(cand \gets $\Call{GetNextCandidate}$) \neq \varnothing$}\label{YKAlgo:GenerateCandidate}
        \For{$i = 1$ \textbf{to} $|D|$}\label{YKAlgo:GetDistancesStart}
        \State $dist_i \gets sdist(D_i, cand)$
        \State place $dist_i$ on $order\_line$
        \EndFor\label{YKAlgo:GetDistancesEnd}
        \State $IG, SD \gets CheckCandidate(order\_line)$\label{YKAlgo:CheckCandidate}
        \If{$IG > bsf\_InfoGain$}
        \State $bsf\_InfoGain \gets IG$
        \State $bsf\_SplitDist \gets SD$
        \State $bsfShapelet \gets cand$
        \State $bsf\_OrderLine \gets order\_line$
        \EndIf
        \EndWhile
        \EndFor
        \State \textbf{return} $bsfShapelet,\ bsf\_SplitDist,\ bsf\_OrderLine$
    \end{algorithmic}
\end{algorithm}


\subsection{Speed-up Strategies}\label{section:speed_up_techniques}
A number of techniques have been proposed to speed-up the shapelet discovery process.
The authors of the YK-Shapelets algorithm proposed early abandoning distance calculations and early candidate pruning using an upper-bound on the information gain 
and reported a speed-up of three orders of magnitude compared to the brute force approach \cite{Ye2009Shapelets}.
The \textit{Logical-Shapelets} algorithm reduces computational costs by reusing previously calculated distances and pruning candidates using the triangular inequality \cite{Mueen2011Logical}.
It can reduce the computational complexity to $O(k^2m^3)$, however, caching the computations imposes a memory requirement on the order of $O(km^2)$ which limits the use of this algorithm for large datasets in memory constrained environments.

The \textit{Fast-Shapelets} algorithm reduces the dimensionality of the data using SAX \cite{Lin2007SAX} and then performs a random projection based shapelet discovery using this lower dimensional data \cite{Rakthanmanon2013FastShapelets}.
It uses a heuristic approach and provides a huge reduction in computational costs but requires tuning a number of parameters according to the dataset characteristics or performance requirements.


The \textit{Random-Shapelets} algorithm performs a uniform random sampling of candidates to reduce the number of evaluated candidates \cite{Renard2015RandomShapelets}.
The YK-Shapelets algorithm generates candidates with a unit step size giving an almost complete overlap to subsequent candidates while uniform sampling effectively increases the step size between subsequent candidates.
Figure \ref{fig:CandidatesExample} shows an example of the first ten candidates generated using the YK-Shapelets and Random-Shapelets algorithms.
The candidates generated using the YK-Shapelets approach have a very high overlap and only cover a small section of the time series.
The candidates generated using random sampling have less overlap and also cover a greater section of the time series.
This reduces the number of prospective candidates but the classification accuracy does not deteriorate because the shape of the time series is still covered by the candidates.
Using an $n\%$ sampling reduces the number of possible shapelet candidates to:
\begin{equation*}
\frac{n}{100}\sum_{l = min}^{max}\sum_{i = 1}^{k}(|T_i|-l+1)
\end{equation*}
This expression provides a precise number of candidates, however, the actual number of evaluated candidates can slightly vary because of the random sampling process.

\begin{figure}[tb]
    \centering
    \includegraphics[width=\columnwidth]{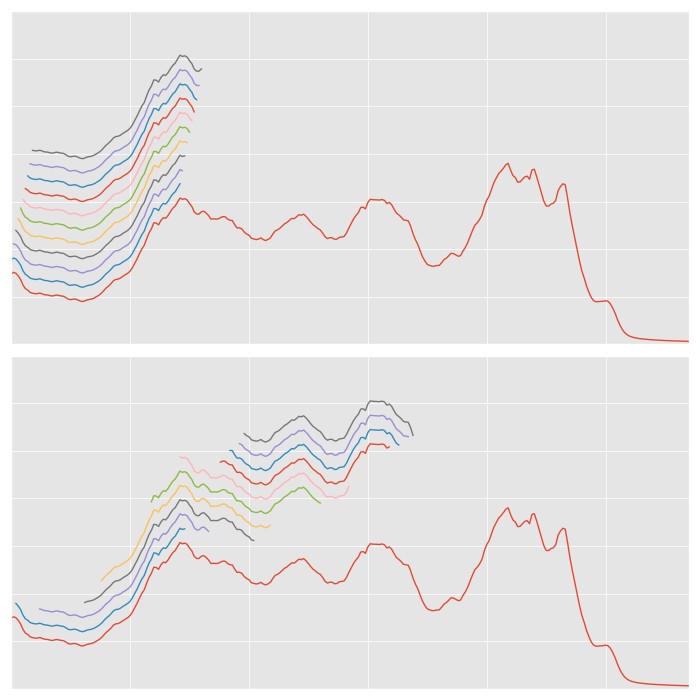}
    \caption{The first ten shapelet candidates generated using: (top) the unit step approach of YK-Shapelets, and (bottom) randomly skipping time points using the Random-Shapelets approach.}
    \label{fig:CandidatesExample}
\end{figure}

A recently published approach called Generalized Random Shapelet Forests (gRSF) also employs ensembles and a randomized candidate sampling based shapelet discovery process for improved classification accuracy and reduced runtime \cite{Karlsson2016gRSF}.
Each ensemble member is trained using a set of instances sampled from the training set with replacement and performing shapelet discovery on this sampled data using a candidate sampling process strikingly similar to the Random-Shapelets approach.
At each node, the gRSF algorithm samples a constant number of candidates proportional to the time series length, while the Random-Shapelets approach samples a percentage of the candidates.

\subsection{Ensemble Methods}\label{section:ensembles_intro}
Ensemble methods are based on the idea of ``combining'' the opinions of different ``experts'' to obtain a decision about a given problem.
Members of an ensemble can have a different view of the data, or they can use different features for making their decision, or they can be totally different algorithms.
This provides diversity in the ensemble decisions, which often makes them more accurate than individual models.
Several studies have shown that ensembles can often be more effective than individual machine learning models.

Due to the variety of proposed combinations in the literature regarding ensembles, it is possible to experiment with a few options.
%
A basic ensemble of classifiers can be obtained by combining multiple diverse models all trained using the same data.
Another approach for constructing ensembles is that of Bagging \cite{Breiman1996bagging} which trains $N$ models, each with a different bootstrap of data such that $|D|$ instances are sampled with replacement from the original dataset.
This introduces a diversification effect and a duplication of instances also allows individual models to be focused on the duplicated instances.
Algorithm \ref{algo:Bagging} provides the pseudo-code for bagging.
Another approach called Boosting \cite{freund1995desicion} relies on weighted instances.
All training instances are assigned equal weights so that the weights' sum equals one.
A model is trained and a classification of training data using this model identifies the misclassified instances whose weights are increased.
Next, the weights of all instances are normalized to keep the sum of weights equal to one.
This, in turn, decreases the weights of correctly classified instances and provides emphasis on the misclassified instances in the next iteration and in many cases leads to an improved overall accuracy of the ensemble.
Algorithm \ref{algo:Boosting} provides the pseudo-code for boosting.

\begin{algorithm}[tb]
    \caption{Bagging Ensemble ($D, N$)}
    \label{algo:Bagging}
    \begin{algorithmic}[1]
        \State $M \gets \varnothing$
        \For{$n=1$ \textbf{to} $N$}
        \State $D_n \gets$ sample $|D|$ instances from $D$ with replacement
        \State $M_n \gets$ train classifier on $D_n$
        \State $M\gets M \cup M_n$
        \EndFor
        \State \textbf{return} $M$
    \end{algorithmic}
\end{algorithm}

\begin{algorithm}[tb]
    \caption{Boosting Ensemble ($D, N$)}
    \label{algo:Boosting}
    \begin{algorithmic}[1]
        \State $M \gets \varnothing$
        \State $w_{1i} \gets \frac{1}{|D|}$, $\forall x_i \in D$
        \For{$n=1$ to $N$}
        \State $M_n \gets$ train classifier on $D$ with $w_{n}$
        \State calculate weighted error $\epsilon_t$
        \If{$\epsilon_n \geq 0.5$}
        \State $N \gets n-1$
        \State $BREAK$
        \Else
        \State $\alpha_n \gets \frac{1}{2} \times ln \frac{1-\epsilon_n}{\epsilon_n}$
        \State $w_{(n+1)}^i \gets \frac{w_{n}^i}{2 \epsilon_n}$, $\forall$ misclassified $x_i \in D$
        \State $w_{(n+1)}^j \gets w_{n}^j$, $\forall$ correctly classified $x_j \in D$
        \State normalize $w_{(n+1)}$
        \EndIf
        \State $M\gets M \cup M_n$
        \EndFor
        \State \textbf{return} $M(x) = \sum_{t=1}^{N}\alpha_n M_n$
    \end{algorithmic}
\end{algorithm}

\section{Proposed Method}\label{section:algo}
In this section we describe the proposed method, which is a combination of the Random-Shapelets algorithm and standard ensemble approaches.
The Random-Shapelets algorithm is computationally less expensive than the YK-Shapelets algorithm, however, it generates models with slightly variable results.
This variability makes the Random-Shapelets algorithm a diverse algorithm and can be used to our advantage.
The lesser computational cost and inherent randomization of the Random-Shapelets algorithm makes it a prime candidate for incorporation in an ensemble.
Therefore, we can use the Random-Shapelets algorithm as the base classifier in an ensemble learning approach for the time series classification problem.
This combines the strengths of ensemble learning with the efficient but non-exact shapelet discovery process of Random-Shapelets to provide a highly cost effective alternative to the exact YK-Shapelets approach.
Another benefit of choosing Random-Shapelets as the base classifier is that it requires a single parameter, i.e.\ the sampling ratio, which allows to reduce the number of evaluated candidates and directly corresponds to the amount of computation we are willing to spend for finding the shapelets.
Using a small value for the sampling ratio provides speed-up while a higher value provides results which are more consistent with those of the brute force approach.

The method proposed in this paper, called Ensembles of Random Shapelets (EnRS), trains a set of shapelet based classifiers.
The precise working of the approach can be described as follows.
Depending on its incarnation, we use either bagging (Algorithm \ref{algo:Bagging}) or boosting (Algorithm \ref{algo:Boosting}) for training the individual models in EnRS-Bagging or EnRS-Boosting, respectively.
In the EnRS-Bagging approach, different bootstraps of data are used to train Random-Shapelets models.
This incorporates dual randomization in the overall process.
The input for each model is randomized while the shapelet discovery process is already randomized.
This also provides instance duplication which allows the shapelet discovery process to quickly separate duplicated instances and then efficiently perform a search for shapelets in the other instances.
This also provides an effective speed-up for the discovery process.
In the EnRS-Boosting approach, instances are weighted and each iteration trains a model more focused on the misclassified instances from the previous iterations.
This approach also modifies the calculation of information gain to use the weights instead of the class counts in the current split.
In addition to the EnRS-Bagging and EnRS-Boosting variants, we have also included a variant denoted by EnRS, which builds an ensemble of Random-Shapelets based trees with the original training set and does not make use of bootstrap sampling or boosting. It combines multiple Random-Shapelets models and relies only on the diversification provided by the base classifier.

The model generation within the ensemble methods use the decision tree construction from Algorithm \ref{algo:decision-tree}, which in turn uses the shapelet discovery process (Algorithm \ref{algo:YKAlgo}) using the randomized candidate generation approach (Algorithm \ref{algo:GetRandomCandidate}).
In contrast to the YK-Shapelets algorithm, which evaluates all possible shapelet candidates, the Random-Shapelets algorithm adopted in this paper only evaluates a small percentage of the candidates chosen at random.
This is the main difference between the two algorithms and they completely share the rest of the shapelet discovery process.
The candidate sampling is performed while generating shapelet candidates.
The basic procedure is the same, however, a loop skips candidates based on a uniformly distributed random number and the provided selection probability.
The final classification decision for an instance is based on voting.

\begin{algorithm}[tb]
	\caption{GetNextCandidate ()}
    \label{algo:GetRandomCandidate}
	\begin{algorithmic}[1]
		\State $cand \gets \varnothing$
		\While{$CandidatesAvailable() = True$}
		\State update $currPos, currInstance, currLen$
		\State $rand\_num \sim U(0,1)$\label{RandomShapeletAlgo:DrawRandom}
		\If{$rand\_num < ratio$}
		\State $cand \gets $ create candidate
		\State BREAK
		\EndIf\label{RandomShapeletAlgo:SelectionDone}
		\EndWhile
		\State \textbf{return} $cand$
	\end{algorithmic}
\end{algorithm}

\subsection{Optimizations}
The extent of sampling has a direct influence on the overall reduction in the  computational costs achieved by the Random-Shapelets algorithm.
Smaller sampling ratios lead to higher reduction and vice versa.
The authors of the Random-Shapelets algorithm only evaluated the effects of sampling the shapelet candidates, however, the approach can be further optimized by incorporating the different speed up techniques proposed in some other research efforts.
First and foremost, the early abandoning of distance calculations and early candidate pruning approaches introduced in the YK-Shapelets paper can be used in the Random-Shapelets algorithm as well.
Distance calculations are abandoned as soon as the distance between the candidate and the subsequence from the current window exceed the running ``best so far'' value.
The candidates themselves are pruned based on an inexpensive information gain computation.
The idea is to compute an optimistic information gain value after placing each time series instance on the order line to estimate whether such a placement of the remaining instances will provide a better information gain than the ``best so far'' information gain.
If the optimistic information gain is better than the best so far value, then the current candidate can provide a better order line and, therefore, it is potentially a better candidate than the current best shapelet.
So we continue the evaluation of the remaining time series instances in the dataset.
However, if the optimistic ``information gain'' is less than the best so far value, then any further processing of the current candidate can not lead to a better result so the evaluation of the candidate shapelet can be abandoned altogether.
These speed-up techniques provide an inexpensive but highly effective way of pruning distance calculations and the shapelet candidates.

\subsection{Normalization}
Normalizing subsequences before distance calculations provides better overall accuracy for the shapelet discovery algorithm because time series similarity benefits from scale and offset invariance.
Therefore, we need to z-normalize the subsequences before distance calculations.
This requires the calculation of mean and standard deviation values for each subsequence prior to the distance calculation.
Calculation of these values makes up the majority of the computation required for normalizing the subsequences and makes the computational costs untenable.
Using a ``summary statistics'' \cite{Sakurai2005BRAID} based approach can drastically reduce the amount of computation required to calculate the mean and standard deviation of a given subsequence.
The ``summary statistics'' for each time series instance in the training set are calculated at the time of loading the dataset, while for the shapelet candidates, they are calculated at the time of shapelet candidate generation.
Subsequently, a simple procedure can provide the mean and standard deviation value in almost constant time and drastically reduce the cost of z-normalizing the sequences before distance calculations.
This approach for z-normalization of subsequences has also been incorporated in our implementation to improve the overall accuracy of the models and also for reduced computational overhead.

\section{Experimental Design}\label{section:experimental-design}
The main goal of our experimental evaluation is twofold.
We want to evaluate whether 
(1) the proposed approach provides better or at par classification accuracy compared to the YK-Shapelets approach and
(2) whether it requires less computational effort as compared to the exact method.
Therefore, an extensive set of experiments has been carried out for the  evaluation and comparison of the different approaches.
We have set the classification accuracy and runtime of the YK-Shapelets approach as the baseline.
We have also compared the results for the Fast-Shapelets algorithm \cite{Rakthanmanon2013FastShapelets} and the gRSF algorithm \cite{Karlsson2016gRSF}.

The YK-Shapelets, Random-Shapelets and the three ensemble approaches have been implemented in Java using a consistent program structure so that no algorithm gets an undue bias.
For boosting only, the shapelet discovery procedure uses weighted instances and the weights of the instances are used instead of instance counts for calculating the entropy and information gain of the datasets and splits.
Both the YK-Shapelets and Random-Shapelets algorithms share the core functionality and implementation so we can effectively compare the running times for the different approaches and determine the obtained speed up.
Our implementation of the different algorithms is available from the accompanying web page for the paper.\footnote{ \url{https://dx.doi.org/10.6084/m9.figshare.4299521}}
The Fast-Shapelets Java implementation was obtained from the UEA Time Series Repository.\footnote{ \url{http://www.timeseriesclassification.com}}
The gRSF implementation was obtained from the supporting web page for the paper.\footnote{ \url{http://people.dsv.su.se/~isak-kar/grsf/}}

\subsection{Datasets}\label{section:datasets}
The empirical evaluation has been carried out on \totalDatasets{} datasets publicly available from the UCR Time Series Archive.\footnote{ \url{http://www.cs.ucr.edu/~eamonn/time_series_data/}}
The datasets belong to various fields including ECG readings, image outlines, motion capture data, sensor readings, spectral analyses and synthetically generated data.
The problems addressed in these datasets range from binary to multi-class problems.
The original training and testing splits are used as such, using the training splits to train the classifiers and reporting the training time while using the testing set to report classification accuracy.

\subsection{Experiments}\label{section:experiments}
We have carried out experiments for the YK-Shapelets, Fast-Shapelets, Random-Shapelets and ensembles of Random-Shapelets based classifiers.
Since YK-Shapelets is an exact method and its classification accuracy remains the same over different runs provided the candidate length parameters are kept the same, each dataset is evaluated once.
For all the other algorithms, each dataset is evaluated 100 times with each algorithm and the mean and standard deviation of the achieved classification accuracy are reported.
The number of classification models per ensemble is set to ten for each variant and voting is used for the final decision.
We use fully grown decision tree models in all our experiments for all the algorithms.

\subsection{Parameter Settings}\label{section:params}
The main parameters required for the shapelet discovery process are $minLen$ and $maxLen$, which define the range of possible shapelet candidate lengths.
The shapelet discovery process searches for candidates in all possible window sizes between the provided minimum and maximum length sizes.
For example, if $minLen=10$ and $maxLen=20$, then the shapelet discovery process will search for shapelets in all window sizes starting from 10 and ending at 20.
Therefore, setting these values to the extreme cases, where $minLen=1$ and $maxLen=m$, where $m$ is the time series instance length, makes the algorithm search over the entire candidate set.
Another approach is to set these parameters based on some assumptions about the possible shapelet lengths.
However, setting these parameters incorrectly can be detrimental to the shapelet discovery process.
Setting the parameters to a very small window can cause the shapelet discovery process to miss important features because they are not covered in the search window while setting the window to a very large size can cause suboptimal feature selection.
A third approach is to use a parameter optimization phase before creating the complete classification model.

The experiments were executed using two main approaches.
For the first approach, instead of setting the parameter values to the extreme cases, or making any assumptions about the possible shapelet lengths, we take a cautious approach and set the parameters to a constant fraction of the time series length for all datasets.
For all experiments, we used $minLen = \lceil0.25 \times m\rceil$ and $maxLen = \lfloor0.67 \times m\rfloor$, where $m$ is the length of time series.
This allows to cover more than 66\% of the time series length at the start of the discovery process and narrows the search up to just a quarter of the time series length.
For our second approach, we used a parameter optimization phase to search for the best shapelet candidate lengths for each dataset and then performed the experiments using these learned parameters.
The parameter optimization was performed with only the training split of the datasets.

The Random-Shapelets algorithm evaluates a small fraction of all the possible candidates in the specified $minLen$ and $maxLen$ range.
This fraction of candidates is determined by the sampling ratio.
All the experiments involving the Random-Shapelets algorithm have been performed with a 1\% sampling ratio.
This includes the experiments for evaluating the Random-Shapelets algorithm itself and the variants of ensembles of Random-Shapelets.

\section{Results}\label{section:results}
We will evaluate the competing algorithms on the basis of classification accuracy and the required computational cost.
We used \totalDatasets{} different datasets for a thorough evaluation of the algorithms.
All the experiments have been carried out on a High Performance Cluster.
The maximum allowed time for evaluating a dataset with any given algorithm was set at 10 days.
If the experiment for a dataset did not complete in that time frame, we report it as ``DNF''.
We will summarize the results in this section.\footnote{Detailed results can be obtained from \url{https://dx.doi.org/10.6084/m9.figshare.4299479}}

\subsection{Classification Accuracy}\label{section:accuracy}
In our experiments, the ensembles of Random-Shapelets classifiers consistently outperformed the other algorithms and provided better classification accuracy.
This observation is particularly interesting because the individual models in the ensembles were using just 1\% of the possible candidates, uniformly sampled from the set of all possible candidates.

Figures \ref{subfig:fixed_params} and \ref{subfig:tuned_params} show the critical differences diagram for the classification accuracies of the individual algorithms for a $p=0.05$ significance level.
The ensembles outperform the other algorithms.
The large difference between the average ranks of the ensembles and other algorithms, especially YK-Shapelets, shows that the average improvement in the classification accuracy is also significant.
For many datasets, the improvement in classification accuracy was as high as 20\% when using ensembles as compared to the classification accuracy achieved using the YK-Shapelets algorithm.
The total number of wins for ensembles against the other algorithms is \fixedWins{} and \tunedWins{} for fixed and tuned parameters respectively.
A very peculiar thing to note is that the standalone Random-Shapelets algorithm can also outperform the YK-Shapelets algorithm with a significant difference.
This happens because the YK-Shapelets model overfits the data whereas the Random-Shapelets model can better generalize on the test set.
The results of experiments performed with fixed parameter settings are provided in Table \ref{table:fixedParams} while the results of experiments using parameter optimization are provided in Table \ref{table:tunedParams}.

\begin{figure}[t]
    \centering
    \subfloat[Fixed parameters\label{subfig:fixed_params}]{
        \includegraphics[width=0.97\columnwidth]{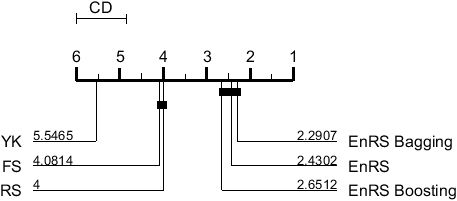}
    }
    
    \subfloat[Tuned parameters\label{subfig:tuned_params}]{
        \includegraphics[width=0.97\columnwidth]{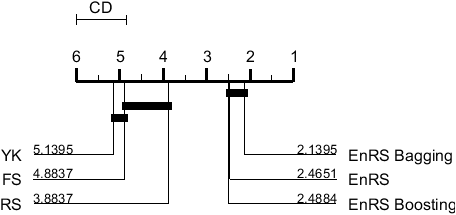}
    }
    \caption{Average ranks for YK-Shapelets (YK), Fast-Shapelets (FS), Random-Shapelets (RS) and Ensembles of Random-Shapelets classifiers (Bagging: EnRS Bagging, Boosting: EnRS Boosting and Simple Combination: EnRS) for the \totalDatasets{} datasets. Groups of classifiers not significantly different (at $p=0.05$) are connected.}
    \label{fig:ClassificationCD}
\end{figure}

\begin{table*}
	\renewcommand{\arraystretch}{1.2}
    \centering
    \caption{Classification accuracy achieved (mean and standard deviation reported) using a fixed shapelet candidate range $minLen = \lceil0.25 \times m\rceil$ and $maxLen = \lfloor0.67 \times m\rfloor$}
    \label{table:fixedParams}
    \rowcolors{2}{white}{gray!15}
    \begin{adjustbox}{max width=\textwidth}
    \begin{tabular}{l|c|c|c|c|c|c}
    \begin{tabular}[]{@{}c@{}}\\Datasets\end{tabular}&
    \multicolumn{1}{ |c| }{\begin{tabular}[c]{@{}c@{}}YK-\\Shapelets\end{tabular}}&
    \multicolumn{1}{ |c| }{\begin{tabular}[c]{@{}c@{}}Fast-\\Shapelets\end{tabular}}&
    \multicolumn{1}{ |c| }{\begin{tabular}[c]{@{}c@{}}Random-\\Shapelets\end{tabular}}&
    \multicolumn{1}{ |c| }{\begin{tabular}[c]{@{}c@{}}EnRS\\ \end{tabular}}&
    \multicolumn{1}{ |c| }{\begin{tabular}[c]{@{}c@{}}EnRS\\Bagging\end{tabular}}&
    \multicolumn{1}{ |c  }{\begin{tabular}[c]{@{}c@{}}EnRS\\Boosting\end{tabular}}\\
    
    \hline
    \hline
	
    50words                      &  47.91 $\pm$ 0.00 &  50.22 $\pm$ 1.25 &  48.87 $\pm$ 2.21 &  68.71 $\pm$ 1.31 &  65.45 $\pm$ 1.49 &  \textbf{69.10 $\pm$ 1.36}\\
    Adiac                        &  43.99 $\pm$ 0.00 &  44.96 $\pm$ 1.78 &  48.37 $\pm$ 2.62 &  64.24 $\pm$ 1.45 &  61.32 $\pm$ 1.58 &  \textbf{64.51 $\pm$ 1.56}\\
    ArrowHead                    &  57.71 $\pm$ 0.00 &  61.26 $\pm$ 5.51 &  63.89 $\pm$ 3.81 &  65.83 $\pm$ 2.18 &  \textbf{67.83 $\pm$ 3.31} &  65.94 $\pm$ 2.15\\
    Beef                         &  56.67 $\pm$ 0.00 &  59.67 $\pm$ 5.08 &  54.20 $\pm$ 6.15 &  60.43 $\pm$ 4.98 &  60.37 $\pm$ 6.75 &  \textbf{60.60 $\pm$ 4.70}\\
    BeetleFly                    &  65.00 $\pm$ 0.00 &  65.00 $\pm$ 0.00 &  69.60 $\pm$ 6.46 &  77.15 $\pm$ 4.40 &  70.85 $\pm$ 9.02 &  \textbf{77.30 $\pm$ 4.11}\\
    BirdChicken                  &  70.00 $\pm$ 0.00 &  59.00 $\pm$ 7.75 &  81.20 $\pm$11.04 &  89.40 $\pm$ 3.35 &  78.10 $\pm$ 5.49 &  \textbf{89.45 $\pm$ 2.35}\\
    CBF                          &  92.78 $\pm$ 0.00 &  \textbf{94.23 $\pm$ 1.25} &  89.14 $\pm$ 4.44 &  93.04 $\pm$ 1.93 &  92.44 $\pm$ 2.34 &  92.73 $\pm$ 1.89\\
    Car                          &  76.67 $\pm$ 0.00 &  71.67 $\pm$ 3.51 &  76.45 $\pm$ 3.60 &  \textbf{76.98 $\pm$ 2.07} &  76.13 $\pm$ 4.53 &  76.95 $\pm$ 1.73\\
    Coffee                       &  82.14 $\pm$ 0.00 &  92.50 $\pm$ 1.13 &  90.62 $\pm$ 3.63 &  \textbf{94.93 $\pm$ 1.84} &  92.39 $\pm$ 3.24 &  94.61 $\pm$ 2.18\\
    DiatomSizeReduction          &  71.24 $\pm$ 0.00 &  \textbf{88.46 $\pm$ 2.50} &  79.41 $\pm$ 6.32 &  83.81 $\pm$ 5.87 &  87.70 $\pm$ 5.57 &  83.53 $\pm$ 5.07\\
    DistalPhalanxOutlineAgeGroup &  72.00 $\pm$ 0.00 &  66.91 $\pm$ 3.16 &  76.32 $\pm$ 3.05 &  82.48 $\pm$ 1.01 &  82.73 $\pm$ 1.19 &  \textbf{82.82 $\pm$ 1.01}\\
    DistalPhalanxOutlineCorrect  &  72.83 $\pm$ 0.00 &  73.44 $\pm$ 2.65 &  72.65 $\pm$ 2.47 &  78.18 $\pm$ 0.96 &  \textbf{79.27 $\pm$ 1.31} &  78.99 $\pm$ 1.08\\
    DistalPhalanxTW              &  71.00 $\pm$ 0.00 &  63.96 $\pm$ 3.45 &  68.83 $\pm$ 2.48 &  74.10 $\pm$ 1.26 &  \textbf{78.00 $\pm$ 1.17} &  74.16 $\pm$ 1.34\\
    ECG200                       &  80.00 $\pm$ 0.00 &  76.90 $\pm$ 3.14 &  76.91 $\pm$ 4.42 &  \textbf{81.68 $\pm$ 2.92} &  81.42 $\pm$ 2.60 &  81.53 $\pm$ 2.81\\
    ECGFiveDays                  &  96.17 $\pm$ 0.00 &  \textbf{99.70 $\pm$ 0.30} &  97.95 $\pm$ 1.55 &  98.81 $\pm$ 0.62 &  98.36 $\pm$ 1.19 &  98.81 $\pm$ 0.63\\
    FISH                         &  DNF              &  73.94 $\pm$ 3.16 &  78.24 $\pm$ 3.18 &  86.87 $\pm$ 1.95 &  \textbf{88.11 $\pm$ 1.79} &  87.23 $\pm$ 1.80\\
    FaceAll                      &  56.57 $\pm$ 0.00 &  61.64 $\pm$ 1.52 &  61.65 $\pm$ 1.81 &  72.14 $\pm$ 0.55 &  71.75 $\pm$ 0.97 &  \textbf{72.21 $\pm$ 0.60}\\
    FaceFour                     &  79.55 $\pm$ 0.00 &  \textbf{89.77 $\pm$ 1.42} &  80.53 $\pm$ 7.79 &  89.36 $\pm$ 3.38 &  88.64 $\pm$ 4.39 &  89.23 $\pm$ 3.41\\
    FacesUCR                     &  64.00 $\pm$ 0.00 &  68.36 $\pm$ 2.63 &  66.77 $\pm$ 2.86 &  85.66 $\pm$ 0.95 &  83.28 $\pm$ 1.32 &  \textbf{85.79 $\pm$ 1.06}\\
    Gun\_Point                   &  93.33 $\pm$ 0.00 &  93.87 $\pm$ 3.28 &  93.93 $\pm$ 2.19 &  97.10 $\pm$ 1.33 &  96.71 $\pm$ 1.41 &  \textbf{97.22 $\pm$ 1.23}\\
    Herring                      &  53.12 $\pm$ 0.00 &  \textbf{62.81 $\pm$ 5.65} &  55.92 $\pm$ 8.20 &  54.41 $\pm$ 5.35 &  58.23 $\pm$ 5.60 &  54.12 $\pm$ 4.42\\
    InsectWingbeatSound          &  48.64 $\pm$ 0.00 &  48.85 $\pm$ 1.92 &  47.23 $\pm$ 1.55 &  \textbf{57.21 $\pm$ 0.91} &  57.12 $\pm$ 1.29 &  57.10 $\pm$ 0.98\\
    ItalyPowerDemand             &  94.85 $\pm$ 0.00 &  85.96 $\pm$ 4.45 &  90.53 $\pm$ 3.69 &  \textbf{95.25 $\pm$ 0.59} &  94.83 $\pm$ 0.83 &  95.19 $\pm$ 0.51\\
    Lighting2                    &  \textbf{75.41 $\pm$ 0.00} &  68.52 $\pm$ 4.36 &  68.07 $\pm$ 7.49 &  75.10 $\pm$ 3.58 &  67.80 $\pm$ 5.04 &  75.30 $\pm$ 3.52\\
    Lighting7                    &  54.79 $\pm$ 0.00 &  57.53 $\pm$ 3.93 &  58.85 $\pm$ 4.87 &  65.19 $\pm$ 2.57 &  \textbf{67.59 $\pm$ 3.43} &  65.45 $\pm$ 2.78\\
    MALLAT                       &  82.26 $\pm$ 0.00 &  89.47 $\pm$ 2.28 &  87.36 $\pm$ 2.16 &  89.13 $\pm$ 0.65 &  \textbf{90.77 $\pm$ 1.96} &  89.19 $\pm$ 0.60\\
    Meat                         &  83.33 $\pm$ 0.00 &  78.83 $\pm$ 1.58 &  88.20 $\pm$ 2.57 &  \textbf{88.70 $\pm$ 1.78} &  85.88 $\pm$ 2.85 &  88.40 $\pm$ 1.64\\
    MiddlePhalanxOutlineAgeGroup &  73.50 $\pm$ 0.00 &  53.18 $\pm$ 4.17 &  73.37 $\pm$ 1.74 &  76.26 $\pm$ 0.95 &  \textbf{76.52 $\pm$ 1.21} &  76.21 $\pm$ 1.11\\
    MiddlePhalanxOutlineCorrect  &  73.83 $\pm$ 0.00 &  71.10 $\pm$ 3.36 &  66.63 $\pm$ 5.06 &  \textbf{74.54 $\pm$ 2.36} &  71.13 $\pm$ 4.13 &  72.42 $\pm$ 2.86\\
    MiddlePhalanxTW              &  54.14 $\pm$ 0.00 &  51.62 $\pm$ 3.51 &  55.62 $\pm$ 1.71 &  57.99 $\pm$ 1.25 &  \textbf{59.19 $\pm$ 1.27} &  57.94 $\pm$ 1.15\\
    MoteStrain                   &  79.55 $\pm$ 0.00 &  70.58 $\pm$ 1.41 &  77.83 $\pm$ 4.64 &  85.10 $\pm$ 2.03 &  85.00 $\pm$ 2.78 &  \textbf{85.47 $\pm$ 1.65}\\
    OliveOil                     &  73.33 $\pm$ 0.00 &  73.33 $\pm$ 0.00 &  79.56 $\pm$ 4.58 &  81.43 $\pm$ 3.80 &  \textbf{83.80 $\pm$ 5.55} &  81.50 $\pm$ 3.71\\
    Plane                        &  93.33 $\pm$ 0.00 &  \textbf{96.95 $\pm$ 2.83} &  94.39 $\pm$ 2.25 &  96.25 $\pm$ 0.57 &  96.83 $\pm$ 1.26 &  96.10 $\pm$ 0.51\\
    ProximalPhalanxTW            &  68.75 $\pm$ 0.00 &  72.29 $\pm$ 2.62 &  68.90 $\pm$ 2.20 &  74.81 $\pm$ 1.32 &  \textbf{78.27 $\pm$ 1.09} &  76.88 $\pm$ 1.53\\
    ShapeletSim                  &  46.67 $\pm$ 0.00 &  60.39 $\pm$11.86 &  78.09 $\pm$11.72 &  83.71 $\pm$ 3.03 &  58.69 $\pm$ 5.95 &  \textbf{83.71 $\pm$ 2.80}\\
    SonyAIBORobotSurface         &  \textbf{88.02 $\pm$ 0.00} &  68.55 $\pm$ 0.00 &  83.12 $\pm$ 5.47 &  86.45 $\pm$ 1.84 &  84.14 $\pm$ 3.29 &  86.24 $\pm$ 1.98\\
    SonyAIBORobotSurfaceII       &  83.00 $\pm$ 0.00 &  78.37 $\pm$ 2.02 &  78.29 $\pm$ 5.47 &  \textbf{87.78 $\pm$ 1.68} &  82.77 $\pm$ 3.36 &  87.72 $\pm$ 2.02\\
    SwedishLeaf                  &  73.12 $\pm$ 0.00 &  72.93 $\pm$ 1.73 &  75.51 $\pm$ 2.09 &  \textbf{88.22 $\pm$ 1.00} &  86.57 $\pm$ 1.19 &  88.14 $\pm$ 0.81\\
    Symbols                      &  76.18 $\pm$ 0.00 &  \textbf{83.74 $\pm$ 1.30} &  77.38 $\pm$ 5.26 &  82.71 $\pm$ 1.62 &  83.14 $\pm$ 3.34 &  82.68 $\pm$ 1.46\\
    ToeSegmentation1             &  88.60 $\pm$ 0.00 &  85.04 $\pm$ 2.23 &  90.66 $\pm$ 1.33 &  \textbf{90.72 $\pm$ 0.63} &  89.37 $\pm$ 1.46 &  90.63 $\pm$ 0.58\\
    ToeSegmentation2             &  \textbf{86.92 $\pm$ 0.00} &  84.77 $\pm$ 6.45 &  85.27 $\pm$ 3.07 &  84.51 $\pm$ 1.05 &  77.13 $\pm$ 4.37 &  84.74 $\pm$ 1.17\\
    Trace                        &  97.00 $\pm$ 0.00 &  99.80 $\pm$ 0.63 &  98.77 $\pm$ 1.39 &  99.89 $\pm$ 0.40 &  99.91 $\pm$ 0.32 &  \textbf{99.98 $\pm$ 0.14}\\
    TwoLeadECG                   &  88.50 $\pm$ 0.00 &  92.48 $\pm$ 0.95 &  89.10 $\pm$ 4.29 &  \textbf{93.87 $\pm$ 1.25} &  92.30 $\pm$ 2.30 &  93.80 $\pm$ 1.54\\
    Wine                         &  \textbf{74.07 $\pm$ 0.00} &  59.44 $\pm$ 6.32 &  62.72 $\pm$ 9.74 &  69.53 $\pm$ 8.14 &  67.48 $\pm$ 9.74 &  70.19 $\pm$ 7.40\\
    synthetic\_control           &  88.00 $\pm$ 0.00 &  91.53 $\pm$ 2.20 &  89.61 $\pm$ 2.26 &  95.96 $\pm$ 1.02 &  95.22 $\pm$ 1.16 &  \textbf{96.80 $\pm$ 1.03}\\
                                 &                   &                   &                   &                   &                   & \\
    \hline
    Wins:                        &        \textbf{4} &        \textbf{7} &        \textbf{0} &       \textbf{11} &       \textbf{10} &            \textbf{13}
  \end{tabular}
  \end{adjustbox}
\end{table*}

\begin{table*}
	\renewcommand{\arraystretch}{1.2}
    \centering
    \caption{Classification accuracy achieved (mean and standard deviation reported) using parameter optimization with training split to find optimal shapelet candidate range}
    \label{table:tunedParams}
    \rowcolors{2}{white}{gray!15}
    \begin{adjustbox}{max width=\textwidth}
    \begin{tabular}{l|c|c|c|c|c|c}
    \begin{tabular}[]{@{}c@{}}\\Datasets\end{tabular}&
    \multicolumn{1}{ |c| }{\begin{tabular}[c]{@{}c@{}}YK-\\Shapelets\end{tabular}}&
    \multicolumn{1}{ |c| }{\begin{tabular}[c]{@{}c@{}}Fast-\\Shapelets\end{tabular}}&
    \multicolumn{1}{ |c| }{\begin{tabular}[c]{@{}c@{}}Random-\\Shapelets\end{tabular}}&
    \multicolumn{1}{ |c| }{\begin{tabular}[c]{@{}c@{}}EnRS\\ \end{tabular}}&
    \multicolumn{1}{ |c| }{\begin{tabular}[c]{@{}c@{}}EnRS\\Bagging\end{tabular}}&
    \multicolumn{1}{ |c  }{\begin{tabular}[c]{@{}c@{}}EnRS\\Boosting\end{tabular}}\\
    
    \hline
    \hline
	
    50words                      &  48.13 $\pm$ 0.00 &  47.96 $\pm$ 2.86 &  49.87 $\pm$ 2.19 &  66.35 $\pm$ 1.13 &  62.91 $\pm$ 1.45 &  \textbf{66.58 $\pm$ 1.28}\\
    Adiac                        &  46.04 $\pm$ 0.00 &  52.71 $\pm$ 2.28 &  46.64 $\pm$ 2.34 &  58.73 $\pm$ 1.75 &  56.47 $\pm$ 1.68 &  \textbf{58.76 $\pm$ 1.24}\\
    ArrowHead                    &  \textbf{68.57 $\pm$ 0.00} &  62.97 $\pm$ 2.19 &  68.05 $\pm$ 1.71 &  68.04 $\pm$ 0.78 &  65.86 $\pm$ 2.65 &  68.00 $\pm$ 0.76\\
    Beef                         &  56.67 $\pm$ 0.00 &  51.67 $\pm$ 4.23 &  60.87 $\pm$ 7.35 &  \textbf{74.60 $\pm$ 5.05} &  65.70 $\pm$ 6.43 &  74.23 $\pm$ 5.54\\
    BeetleFly                    &  95.00 $\pm$ 0.00 &  69.50 $\pm$11.41 &  86.38 $\pm$11.68 &  97.00 $\pm$ 3.76 &  91.58 $\pm$ 6.87 &  \textbf{97.58 $\pm$ 2.71}\\
    BirdChicken                  &  \textbf{85.00 $\pm$ 0.00} &  71.00 $\pm$12.87 &  69.09 $\pm$ 9.54 &  78.11 $\pm$ 7.16 &  79.40 $\pm$ 8.39 &  80.05 $\pm$ 7.33\\
    Car                          &  \textbf{85.00 $\pm$ 0.00} &  68.33 $\pm$ 0.00 &  75.18 $\pm$ 5.86 &  79.92 $\pm$ 3.34 &  83.62 $\pm$ 4.02 &  79.58 $\pm$ 3.99\\
    CBF                          &  93.89 $\pm$ 0.00 &  88.10 $\pm$ 8.43 &  86.62 $\pm$ 8.04 &  \textbf{96.85 $\pm$ 0.98} &  92.96 $\pm$ 3.68 &  96.52 $\pm$ 1.13\\
    Coffee                       &  89.29 $\pm$ 0.00 &  \textbf{95.36 $\pm$ 1.73} &  91.67 $\pm$ 3.31 &  92.89 $\pm$ 0.36 &  91.63 $\pm$ 3.03 &  93.00 $\pm$ 0.70\\
    DiatomSizeReduction          &  89.22 $\pm$ 0.00 &  87.19 $\pm$ 0.65 &  85.66 $\pm$ 6.15 &  91.79 $\pm$ 2.57 &  \textbf{93.17 $\pm$ 2.82} &  91.78 $\pm$ 2.22\\
    DistalPhalanxOutlineAgeGroup &  79.00 $\pm$ 0.00 &  67.55 $\pm$ 2.66 &  79.00 $\pm$ 1.63 &  81.24 $\pm$ 1.18 &  \textbf{82.03 $\pm$ 1.31} &  81.36 $\pm$ 1.12\\
    DistalPhalanxOutlineCorrect  &  77.67 $\pm$ 0.00 &  73.93 $\pm$ 1.38 &  73.79 $\pm$ 2.54 &  78.76 $\pm$ 1.14 &  \textbf{79.41 $\pm$ 1.27} &  78.49 $\pm$ 1.16\\
    DistalPhalanxTW              &  73.25 $\pm$ 0.00 &  61.65 $\pm$ 2.80 &  67.75 $\pm$ 2.46 &  72.56 $\pm$ 1.61 &  \textbf{76.08 $\pm$ 1.05} &  72.55 $\pm$ 1.50\\
    ECG200                       &  \textbf{86.00 $\pm$ 0.00} &  82.70 $\pm$ 4.60 &  80.78 $\pm$ 3.91 &  81.78 $\pm$ 2.07 &  82.74 $\pm$ 3.08 &  81.66 $\pm$ 1.97\\
    ECGFiveDays                  &  99.65 $\pm$ 0.00 &  99.18 $\pm$ 1.50 &  98.76 $\pm$ 1.91 &  \textbf{99.91 $\pm$ 0.11} &  99.80 $\pm$ 0.42 &  \textbf{99.90 $\pm$ 0.10}\\
    FaceAll                      &  58.88 $\pm$ 0.00 &  61.85 $\pm$ 1.51 &  62.12 $\pm$ 1.74 &  \textbf{72.90 $\pm$ 0.74} &  72.78 $\pm$ 0.86 &  72.83 $\pm$ 0.67\\
    FaceFour                     &  85.23 $\pm$ 0.00 &  90.80 $\pm$ 2.30 &  83.72 $\pm$ 7.43 &  \textbf{93.67 $\pm$ 2.11} &  92.82 $\pm$ 3.46 &  93.19 $\pm$ 1.74\\
    FacesUCR                     &  65.37 $\pm$ 0.00 &  69.03 $\pm$ 2.09 &  67.38 $\pm$ 2.36 &  85.59 $\pm$ 1.02 &  83.46 $\pm$ 1.16 &  \textbf{85.70 $\pm$ 0.85}\\
    FISH                         &  82.29 $\pm$ 0.00 &  29.49 $\pm$ 2.00 &  79.99 $\pm$ 3.09 &  91.61 $\pm$ 1.57 &  \textbf{93.29 $\pm$ 1.81} &  91.79 $\pm$ 1.41\\
    Gun\_Point                   &  93.33 $\pm$ 0.00 &  93.80 $\pm$ 0.32 &  92.86 $\pm$ 2.19 &  96.65 $\pm$ 1.24 &  \textbf{97.84 $\pm$ 1.05} &  96.65 $\pm$ 1.28\\
    Herring                      &  \textbf{70.31 $\pm$ 0.00} &  51.41 $\pm$ 5.48 &  59.91 $\pm$ 5.63 &  65.08 $\pm$ 4.45 &  63.38 $\pm$ 4.86 &  65.02 $\pm$ 5.16\\
    InsectWingbeatSound          &  50.30 $\pm$ 0.00 &  52.82 $\pm$ 1.89 &  50.09 $\pm$ 1.72 &  58.22 $\pm$ 0.77 &  \textbf{58.31 $\pm$ 1.01} &  58.02 $\pm$ 0.90\\
    ItalyPowerDemand             &  \textbf{95.24 $\pm$ 0.00} &  78.73 $\pm$ 6.60 &  89.91 $\pm$ 3.65 &  95.06 $\pm$ 0.69 &  94.81 $\pm$ 0.71 &  95.14 $\pm$ 0.69\\
    Lighting2                    &  \textbf{78.69 $\pm$ 0.00} &  75.25 $\pm$ 4.54 &  76.44 $\pm$ 3.61 &  75.52 $\pm$ 2.37 &  76.67 $\pm$ 2.89 &  75.98 $\pm$ 2.51\\
    Lighting7                    &  63.01 $\pm$ 0.00 &  57.40 $\pm$ 5.83 &  61.48 $\pm$ 5.20 &  67.66 $\pm$ 3.10 &  \textbf{68.30 $\pm$ 3.28} &  67.67 $\pm$ 2.44\\
    MALLAT                       &  86.40 $\pm$ 0.00 &  87.33 $\pm$ 1.00 &  86.74 $\pm$ 3.27 &  93.23 $\pm$ 1.27 &  92.48 $\pm$ 1.56 &  \textbf{93.45 $\pm$ 1.19}\\
    Meat                         &  86.67 $\pm$ 0.00 &  83.33 $\pm$ 0.00 &  84.72 $\pm$ 4.44 &  86.62 $\pm$ 2.53 &  \textbf{87.02 $\pm$ 2.59} &  86.43 $\pm$ 2.98\\
    MiddlePhalanxOutlineAgeGroup &  74.25 $\pm$ 0.00 &  50.45 $\pm$ 4.68 &  72.67 $\pm$ 1.97 &  75.03 $\pm$ 1.02 &  \textbf{76.42 $\pm$ 1.19} &  75.13 $\pm$ 1.00\\
    MiddlePhalanxOutlineCorrect  &  67.83 $\pm$ 0.00 &  72.08 $\pm$ 1.94 &  69.09 $\pm$ 2.66 &  72.96 $\pm$ 1.04 &  \textbf{73.27 $\pm$ 1.48} &  73.08 $\pm$ 1.19\\
    MiddlePhalanxTW              &  59.40 $\pm$ 0.00 &  52.66 $\pm$ 2.80 &  56.92 $\pm$ 2.28 &  59.18 $\pm$ 1.30 &  \textbf{60.56 $\pm$ 1.23} &  59.01 $\pm$ 1.16\\
    MoteStrain                   &  84.74 $\pm$ 0.00 &  78.83 $\pm$ 1.74 &  79.24 $\pm$ 6.34 &  89.15 $\pm$ 1.60 &  88.82 $\pm$ 2.18 &  \textbf{89.39 $\pm$ 1.73}\\
    OliveOil                     &  86.67 $\pm$ 0.00 &  64.00 $\pm$ 2.11 &  85.88 $\pm$ 3.53 &  87.20 $\pm$ 1.99 &  \textbf{89.57 $\pm$ 3.63} &  88.00 $\pm$ 1.90\\
    Plane                        &  95.24 $\pm$ 0.00 &  97.14 $\pm$ 3.27 &  94.97 $\pm$ 2.62 &  97.49 $\pm$ 1.02 &  \textbf{97.66 $\pm$ 1.22} &  97.25 $\pm$ 1.10\\
    ProximalPhalanxTW            &  72.75 $\pm$ 0.00 &  73.41 $\pm$ 1.72 &  71.55 $\pm$ 2.96 &  77.76 $\pm$ 1.21 &  \textbf{78.83 $\pm$ 1.11} &  77.80 $\pm$ 1.34\\
    ShapeletSim                  &  92.78 $\pm$ 0.00 & \textbf{100.00 $\pm$ 0.00} &  92.51 $\pm$ 3.15 &  96.75 $\pm$ 0.61 &  96.51 $\pm$ 0.85 &  96.69 $\pm$ 0.69\\
    SonyAIBORobotSurface         &  93.18 $\pm$ 0.00 &  93.34 $\pm$ 2.21 &  88.68 $\pm$ 8.51 &  94.36 $\pm$ 2.48 &  90.29 $\pm$ 4.70 &  \textbf{94.50 $\pm$ 2.26}\\
    SonyAIBORobotSurfaceII       &  83.00 $\pm$ 0.00 &  78.91 $\pm$ 0.77 &  80.92 $\pm$ 5.66 &  89.23 $\pm$ 1.81 &  87.42 $\pm$ 2.70 &  \textbf{89.79 $\pm$ 1.75}\\
    SwedishLeaf                  &  72.16 $\pm$ 0.00 &  72.19 $\pm$ 1.31 &  72.07 $\pm$ 1.60 &  83.07 $\pm$ 1.03 &  80.86 $\pm$ 1.16 &  \textbf{83.14 $\pm$ 0.94}\\
    Symbols                      &  88.14 $\pm$ 0.00 &  89.16 $\pm$ 0.29 &  87.31 $\pm$ 4.52 &  90.84 $\pm$ 0.84 &  90.64 $\pm$ 1.59 &  \textbf{91.08 $\pm$ 0.83}\\
    ToeSegmentation1             &  92.54 $\pm$ 0.00 &  93.86 $\pm$ 0.41 &  89.60 $\pm$ 3.59 &  92.90 $\pm$ 1.10 &  \textbf{94.49 $\pm$ 1.31} &  93.11 $\pm$ 1.28\\
    ToeSegmentation2             &  \textbf{90.77 $\pm$ 0.00} &  69.23 $\pm$ 0.00 &  78.80 $\pm$11.10 &  82.36 $\pm$ 5.70 &  90.05 $\pm$ 4.13 &  82.22 $\pm$ 5.20\\
    Trace                        & \textbf{100.00 $\pm$ 0.00} &  98.90 $\pm$ 0.74 &  99.41 $\pm$ 0.96 &  99.99 $\pm$ 0.10 &  99.98 $\pm$ 0.14 &  99.97 $\pm$ 0.17\\
    TwoLeadECG                   &  96.14 $\pm$ 0.00 &  90.67 $\pm$ 3.46 &  90.01 $\pm$ 4.92 &  96.22 $\pm$ 2.29 &  \textbf{97.01 $\pm$ 2.26} &  95.97 $\pm$ 2.94\\
    Wine                         &  \textbf{77.78 $\pm$ 0.00} &  76.30 $\pm$ 0.78 &  63.62 $\pm$ 7.55 &  68.17 $\pm$ 5.82 &  69.44 $\pm$ 6.29 &  67.61 $\pm$ 5.43\\
    synthetic\_control           &  92.00 $\pm$ 0.00 &  89.87 $\pm$ 1.38 &  91.02 $\pm$ 2.01 &  97.12 $\pm$ 0.70 &  95.94 $\pm$ 0.81 &  \textbf{97.44 $\pm$ 0.73}\\
                                 &                   &                   &                   &                   &                   & \\
    \hline
    Wins:                        &            \textbf{10} &             \textbf{2} &             \textbf{0} &             \textbf{5} &            \textbf{17} &            \textbf{11}
  \end{tabular}
  \end{adjustbox}
\end{table*}

\subsection{Run Time}\label{section:runtime}
The training time for shapelet-based classifiers accounts for almost the entire run time of the algorithms because the testing time is negligible compared to the training time.
The time required for training classification models using the different approaches were noted for the training phase using standard Java timing utilities.
The Fast-Shapelets algorithm was the fastest overall followed by the Random-Shapelets algorithm and then the ensembles (Bagging, Simple combination, Boosting) and finally the YK-Shapelets.
The ensembles consistently performed faster than the YK-Shapelets algorithm and could obtain a speed-up of more than an order of magnitude on average.
Tables \ref{table:fixedParams_time} and \ref{table:tunedParams_time} list the average time (in seconds) for evaluating each dataset with the corresponding algorithm using fixed and optimized parameters respectively.

\begin{table*}
 	\renewcommand{\arraystretch}{1.2}
    \centering
    \caption{Average training time (in seconds, calculated over 100 runs) for classification models generated with fixed shapelet candidate range $minLen = \lceil0.25 \times m\rceil$ and $maxLen = \lfloor0.67 \times m\rfloor$}
    \label{table:fixedParams_time}
    \rowcolors{2}{white}{gray!15}
    \begin{adjustbox}{max width=\textwidth}
    \begin{tabular}{l|c|c|c|c|c|c}
    \begin{tabular}[]{@{}c@{}}\\Datasets\end{tabular}&
    \multicolumn{1}{ |c| }{\begin{tabular}[c]{@{}c@{}}YK-\\Shapelets\end{tabular}}&
    \multicolumn{1}{ |c| }{\begin{tabular}[c]{@{}c@{}}Fast-\\Shapelets\end{tabular}}&
    \multicolumn{1}{ |c| }{\begin{tabular}[c]{@{}c@{}}Random-\\Shapelets\end{tabular}}&
    \multicolumn{1}{ |c| }{\begin{tabular}[c]{@{}c@{}}EnRS\\ \end{tabular}}&
    \multicolumn{1}{ |c| }{\begin{tabular}[c]{@{}c@{}}EnRS\\Bagging\end{tabular}}&
    \multicolumn{1}{ |c  }{\begin{tabular}[c]{@{}c@{}}EnRS\\Boosting\end{tabular}}\\
    
    \hline
    \hline
    50words                      &         206963.1 &            \textbf{354.6} &           7292.2 &          80245.2 &          36069.3 &         383343.5\\
    Adiac                        &          30048.9 &             \textbf{61.1} &            251.1 &           2294.5 &           2254.3 &           5416.9\\
    ArrowHead                    &           1266.2 &              \textbf{6.1} &             16.0 &            157.8 &             84.6 &            196.1\\
    Beef                         &          13160.0 &             \textbf{58.9} &            161.0 &           1815.9 &            808.9 &           1263.8\\
    BeetleFly                    &            354.9 &             \textbf{14.3} &             16.4 &            184.1 &             58.5 &            204.4\\
    BirdChicken                  &           2800.9 &             \textbf{11.9} &             51.3 &            559.6 &            126.2 &            500.1\\
    CBF                          &              7.7 &              1.1 &              \textbf{0.7} &              2.5 &              2.5 &              2.9\\
    Car                          &          93741.6 &             \textbf{86.6} &            770.2 &           9130.9 &           4110.1 &           8319.8\\
    Coffee                       &             79.8 &              3.5 &              \textbf{1.2} &              7.9 &              7.0 &              7.9\\
    DiatomSizeReduction          &            130.8 &              4.2 &              \textbf{1.5} &             12.1 &             10.8 &             12.6\\
    DistalPhalanxOutlineAgeGroup &          44571.7 &              \textbf{5.0} &            312.5 &           3032.2 &           2119.4 &           4160.2\\
    DistalPhalanxOutlineCorrect  &         323947.3 &             \textbf{10.1} &           3624.6 &          36136.6 &          15943.4 &          55118.8\\
    DistalPhalanxTW              &          16770.2 &              \textbf{6.1} &            263.7 &           2321.8 &            805.6 &           3810.0\\
    ECG200                       &           2110.2 &              \textbf{1.6} &             28.4 &            286.7 &            163.0 &            409.1\\
    ECGFiveDays                  &             13.7 &              \textbf{0.5} &              \textbf{0.5} &              1.8 &              1.3 &              2.3\\
    FISH                         &              DNF &            \textbf{237.0} &           4903.0 &          48644.1 &          36409.8 &          73980.4\\
    FaceAll                      &         131065.5 &             \textbf{82.8} &           2221.6 &          23796.6 &          19191.5 &          50005.5\\
    FaceFour                     &            375.8 &             13.0 &              \textbf{4.2} &             37.7 &             36.0 &             39.3\\
    FacesUCR                     &           1857.1 &             \textbf{24.3} &           2259.9 &          20392.1 &           2926.5 &          30932.4\\
    Gun\_Point                   &            228.9 &              \textbf{1.7} &              4.4 &             37.6 &             23.1 &             51.2\\
    Herring                      &         110869.5 &            \textbf{103.5} &           1175.2 &          12264.1 &           8888.4 &          15367.2\\
    InsectWingbeatSound          &          52266.9 &            \textbf{104.8} &            685.3 &           6849.7 &           5819.7 &          11580.8\\
    ItalyPowerDemand             &              2.6 &              \textbf{0.1} &              1.1 &              4.4 &              3.0 &              6.7\\
    Lighting2                    &         151289.4 &            \textbf{176.8} &           2253.9 &          23594.6 &          17498.4 &          23070.0\\
    Lighting7                    &           9558.7 &             \textbf{49.6} &            192.7 &           1956.4 &           1173.9 &           2767.7\\
    MALLAT                       &          77363.8 &            \textbf{551.1} &            645.7 &           6259.9 &           4941.0 &           6258.9\\
    Meat                         &           1878.9 &             38.5 &             \textbf{17.2} &            167.7 &            328.7 &            184.3\\
    MiddlePhalanxOutlineAgeGroup &          48434.9 &              \textbf{5.0} &            736.4 &           6879.0 &           2847.3 &           9694.9\\
    MiddlePhalanxOutlineCorrect  &         181188.6 &             \textbf{10.8} &           3148.1 &          29006.8 &          13536.5 &          59238.2\\
    MiddlePhalanxTW              &          26233.9 &              \textbf{6.7} &            456.8 &           4434.8 &           2000.7 &           6645.8\\
    MoteStrain                   &              6.8 &              \textbf{0.1} &              0.7 &              1.9 &              1.4 &              2.6\\
    OliveOil                     &           1233.8 &             \textbf{37.3} &             43.4 &            433.5 &            203.3 &            460.0\\
    Plane                        &            191.5 &              7.0 &              \textbf{3.5} &             23.8 &             20.0 &             34.6\\
    ProximalPhalanxTW            &          33524.1 &              \textbf{5.4} &            349.8 &           3436.3 &           2657.6 &          17464.3\\
    ShapeletSim                  &            815.0 &             \textbf{21.7} &             48.8 &            482.4 &             82.7 &            495.0\\
    SonyAIBORobotSurface         &              1.4 &              \textbf{0.2} &              0.5 &              0.9 &              0.8 &              1.3\\
    SonyAIBORobotSurfaceII       &              3.3 &              \textbf{0.1} &              0.7 &              1.8 &              1.2 &              2.6\\
    SwedishLeaf                  &          83643.9 &             \textbf{52.9} &           1694.3 &          16466.1 &           8717.2 &          35983.7\\
    Symbols                      &            600.6 &             17.3 &              \textbf{6.9} &             58.1 &             59.7 &             60.2\\
    ToeSegmentation1             &            648.4 &              \textbf{5.1} &             11.4 &            101.6 &             63.5 &            107.0\\
    ToeSegmentation2             &            274.7 &              7.9 &              \textbf{5.4} &             46.2 &             25.0 &             48.9\\
    Trace                        &            588.3 &             18.4 &              \textbf{7.2} &             65.2 &             63.7 &             81.1\\
    TwoLeadECG                   &              1.9 &              \textbf{0.1} &              0.5 &              0.9 &              0.9 &              1.4\\
    Wine                         &           3921.7 &              \textbf{6.2} &             74.1 &            741.1 &            421.3 &            929.2\\
    synthetic\_control           &           1521.2 &              \textbf{4.3} &             46.5 &            464.8 &            218.3 &           3473.4\\
                                 &                   &                   &                   &                   &                   & \\
    \hline
    Wins:                        &             \textbf{0} &            \textbf{36} &             \textbf{9} &             \textbf{0} &             \textbf{0} &             \textbf{0}
  \end{tabular}
  \end{adjustbox}
\end{table*}

\begin{table*}
	\renewcommand{\arraystretch}{1.2}
    \centering
    \caption{Average training time (in seconds, calculated over 100 runs) for classification models generated using parameter optimization}
    \label{table:tunedParams_time}
    \rowcolors{2}{white}{gray!15}
    \begin{adjustbox}{max width=\textwidth}
    \begin{tabular}{l|c|c|c|c|c|c}
    \begin{tabular}[]{@{}c@{}}\\Datasets\end{tabular}&
    \multicolumn{1}{ |c| }{\begin{tabular}[c]{@{}c@{}}YK-\\Shapelets\end{tabular}}&
    \multicolumn{1}{ |c| }{\begin{tabular}[c]{@{}c@{}}Fast-\\Shapelets\end{tabular}}&
    \multicolumn{1}{ |c| }{\begin{tabular}[c]{@{}c@{}}Random-\\Shapelets\end{tabular}}&
    \multicolumn{1}{ |c| }{\begin{tabular}[c]{@{}c@{}}EnRS\\ \end{tabular}}&
    \multicolumn{1}{ |c| }{\begin{tabular}[c]{@{}c@{}}EnRS\\Bagging\end{tabular}}&
    \multicolumn{1}{ |c  }{\begin{tabular}[c]{@{}c@{}}EnRS\\Boosting\end{tabular}}\\
    
    \hline
    \hline
    50words                      &   146347.2 &        \textbf{4.5} &     2081.7 &    20023.4 &    10275.2 &   118973.4\\
    Adiac                        &      896.0 &        \textbf{5.8} &       14.8 &      161.0 &       84.8 &      384.5\\
    ArrowHead                    &       47.4 &        \textbf{0.5} &        0.9 &        5.6 &        3.9 &        8.2\\
    Beef                         &     1140.1 &       \textbf{12.3} &       17.6 &      129.1 &       85.4 &      144.5\\
    BeetleFly                    &       25.4 &        0.7 &        \textbf{0.5} &        3.1 &        3.0 &        3.6\\
    BirdChicken                  &       32.0 &        0.9 &        \textbf{0.7} &        3.7 &        3.1 &        3.8\\
    Car                          &    19824.4 &       \textbf{36.6} &      168.5 &     1877.3 &      871.1 &     2438.4\\
    CBF                          &        5.4 &        \textbf{0.6} &        \textbf{0.6} &        2.1 &        1.9 &        2.5\\
    Coffee                       &       19.5 &        1.1 &        \textbf{0.6} &        2.4 &        2.2 &        2.8\\
    DiatomSizeReduction          &        2.4 &        0.4 &        \textbf{0.3} &        0.6 &        0.6 &        0.7\\
    DistalPhalanxOutlineAgeGroup &     5921.0 &        \textbf{2.1} &       82.6 &      909.7 &      640.2 &     1568.9\\
    DistalPhalanxOutlineCorrect  &    15227.5 &        \textbf{0.5} &      157.2 &     1155.1 &      825.5 &     2035.0\\
    DistalPhalanxTW              &    21973.9 &        \textbf{4.2} &      251.6 &     2161.4 &      649.4 &     3569.5\\
    ECG200                       &      126.8 &        \textbf{0.2} &        2.2 &       17.6 &       10.6 &       24.9\\
    ECGFiveDays                  &        4.3 &        \textbf{0.4} &        0.3 &        0.9 &        0.8 &        1.1\\
    FaceAll                      &    26330.3 &       \textbf{91.6} &     2302.2 &    24076.0 &    19678.6 &    46765.0\\
    FaceFour                     &      283.9 &       11.6 &        \textbf{3.1} &       28.4 &       27.0 &       28.7\\
    FacesUCR                     &     5917.6 &       \textbf{23.9} &     1498.9 &    13146.5 &     1923.2 &    26253.1\\
    FISH                         &   290690.7 &        \textbf{0.7} &     1857.7 &    19375.2 &    11529.7 &    39553.2\\
    Gun\_Point                   &       10.3 &        \textbf{0.8} &        1.6 &        9.3 &        3.9 &       12.6\\
    Herring                      &     7683.8 &        \textbf{6.6} &      112.6 &     1197.5 &      605.7 &     1552.6\\
    InsectWingbeatSound          &     1921.3 &        \textbf{8.8} &       26.5 &      252.0 &      200.7 &      476.1\\
    ItalyPowerDemand             &       11.1 &        \textbf{0.1} &        1.0 &        3.7 &        2.7 &        5.7\\
    Lighting2                    &     3133.8 &       \textbf{13.6} &       44.4 &      454.6 &      261.2 &      549.4\\
    Lighting7                    &    11885.9 &       \textbf{37.1} &      152.3 &     1676.5 &      742.1 &     2348.1\\
    MALLAT                       &     7844.5 &      143.2 &       \textbf{61.5} &      599.3 &      390.6 &      702.3\\
    Meat                         &       57.2 &        2.0 &        \textbf{1.7} &       10.4 &       12.7 &       13.0\\
    MiddlePhalanxOutlineAgeGroup &     9866.1 &        \textbf{2.1} &      147.4 &     1345.5 &      763.5 &     2281.0\\
    MiddlePhalanxOutlineCorrect  &    11072.4 &        \textbf{0.8} &      124.5 &     1283.4 &      679.5 &     2334.8\\
    MiddlePhalanxTW              &     1580.5 &        \textbf{0.6} &       16.7 &      170.4 &      100.8 &      257.8\\
    MoteStrain                   &        1.0 &        \textbf{0.1} &        0.7 &        1.5 &        1.2 &        2.1\\
    OliveOil                     &      909.4 &       30.8 &       \textbf{13.7} &      132.7 &      116.1 &      141.4\\
    Plane                        &      136.3 &        6.1 &        \textbf{2.9} &       19.2 &       17.4 &       30.1\\
    ProximalPhalanxTW            &      565.5 &        \textbf{0.5} &       23.1 &      206.7 &      166.4 &      782.5\\
    ShapeletSim                  &      229.2 &        7.0 &        \textbf{2.2} &       17.3 &       15.3 &       18.3\\
    SonyAIBORobotSurface         &        1.4 &        \textbf{0.1} &        0.6 &        1.3 &        1.1 &        1.7\\
    SonyAIBORobotSurfaceII       &        2.7 &        \textbf{0.1} &        0.8 &        1.7 &        1.4 &        2.4\\
    SwedishLeaf                  &     7379.2 &       \textbf{12.2} &      254.6 &     2545.0 &     1771.4 &     5022.7\\
    Symbols                      &        6.6 &        1.5 &        \textbf{0.7} &        1.7 &        1.7 &        2.2\\
    ToeSegmentation1             &       30.4 &        \textbf{2.0} &        2.3 &       14.3 &        6.8 &       17.1\\
    ToeSegmentation2             &      114.1 &        \textbf{4.7} &        5.2 &       43.9 &       12.5 &       47.6\\
    Trace                        &      507.7 &       20.8 &        \textbf{5.4} &       47.0 &       43.9 &       64.4\\
    TwoLeadECG                   &        2.0 &        \textbf{0.1} &        0.5 &        1.1 &        1.1 &        1.4\\
    Wine                         &     3290.9 &        \textbf{3.7} &       52.8 &      554.6 &      252.0 &      740.3\\
    synthetic\_control           &      854.5 &        \textbf{1.4} &       13.4 &      142.3 &       76.9 &     1420.8\\
                                 &                   &                   &                   &                   &                   & \\
    \hline
    Wins:                        &           \textbf{0}  &             \textbf{32} &             \textbf{13} &             \textbf{0} &            \textbf{0} &            \textbf{0}
  \end{tabular}
  \end{adjustbox}
\end{table*}

The results for parameter optimized experiments are shown in Figure \ref{fig:tuned_all}. The figures show the accuracy (left panel) and speed-up (right panel) obtained for each dataset.
The results for YK-Shapelets are plotted as a solid black line denoting the baseline while the box plots show the obtained results for the other algorithms.
Any value to the left of the baseline implies that the YK-Shapelets algorithm performed better than the other algorithm.
The speed-up obtained for the algorithms is plotted by dividing the time taken by YK-Shapelets algorithm by the time taken by the respective algorithm.
Any value to the left of the baseline implies the YK-Shapelets algorithm was faster.
The speed-up is plotted on a logarithmic scale so $10^0$ or 1 means no speed-up while $10^1$ implies a speed-up of one order of magnitude.
The red lines in the box plots show the median values for the observations while the black whiskers show the minimum and maximum values for the observations.
\begin{figure*}[p]
    \centering
    \subfloat[50words\label{subfig:50words}]{\includegraphics[width=\columnwidth]{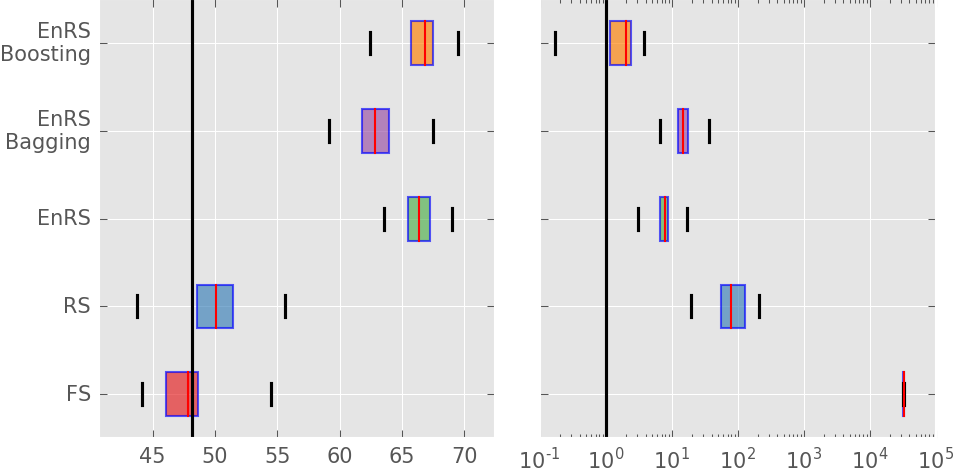}}
    \hspace{0.5cm}
    \subfloat[Adiac\label{subfig:Adiac}]{\includegraphics[width=\columnwidth]{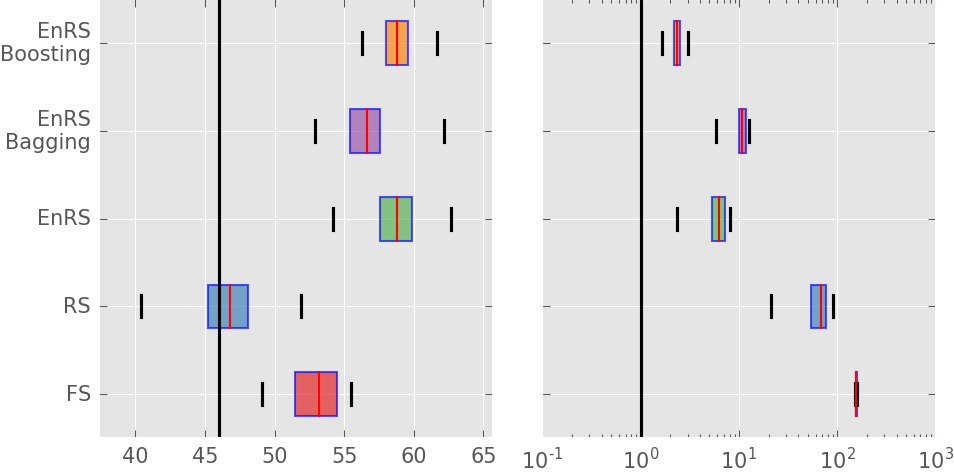}}
    
    \subfloat[ArrowHead\label{subfig:ArrowHead}]{\includegraphics[width=\columnwidth]{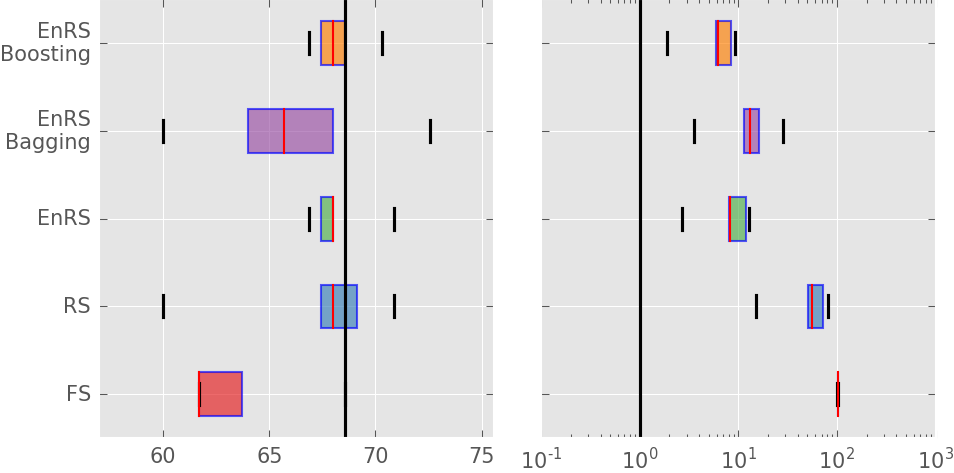}}
    \hspace{0.5cm}
    \subfloat[Beef\label{subfig:Beef}]{\includegraphics[width=\columnwidth]{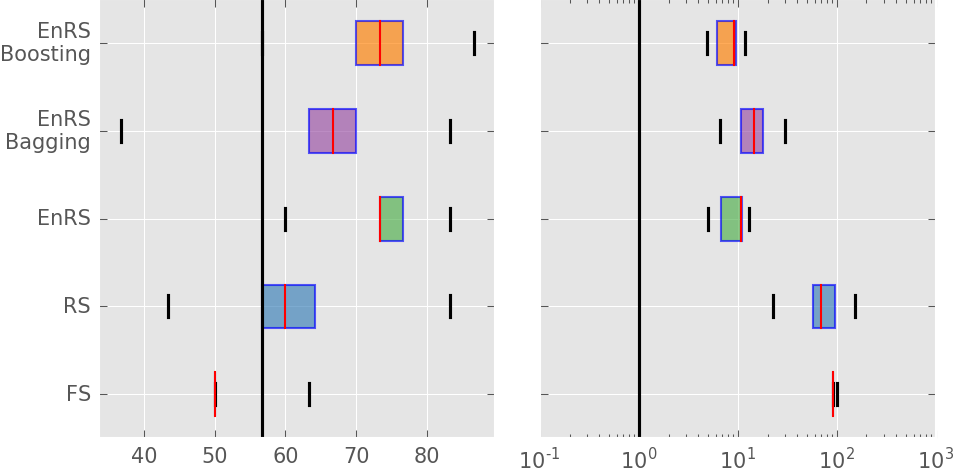}}
    
     \subfloat[BeetleFly\label{subfig:BeetleFly}]{\includegraphics[width=\columnwidth]{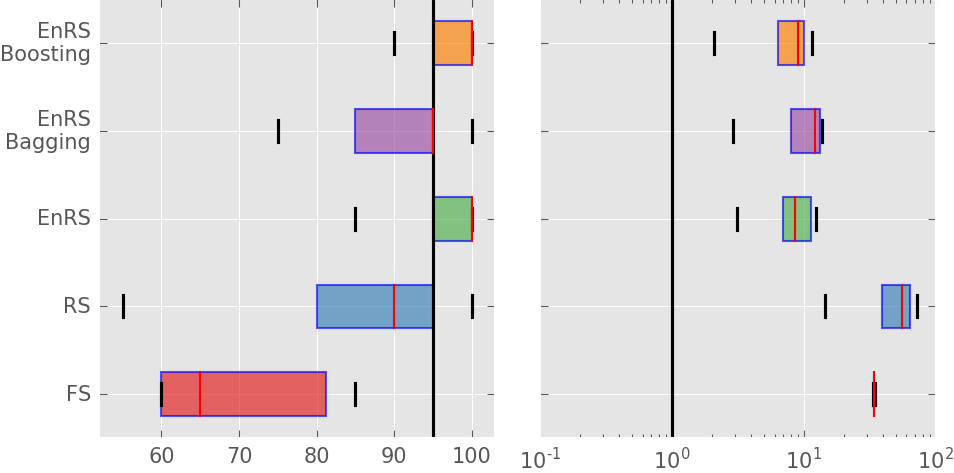}}
    \hspace{0.5cm}
     \subfloat[BirdChicken\label{subfig:BirdChicken}]{\includegraphics[width=\columnwidth]{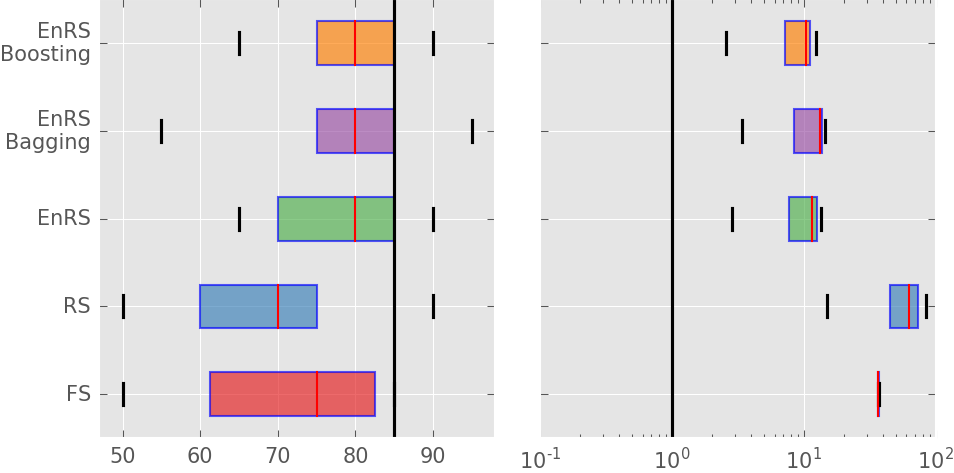}}

     \subfloat[Car\label{subfig:Car}]{\includegraphics[width=\columnwidth]{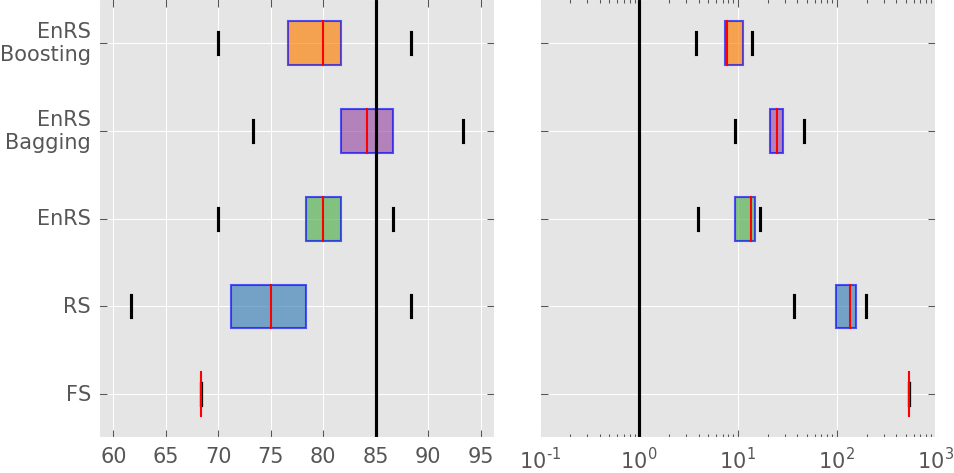}}
    \hspace{0.5cm}
     \subfloat[CBF\label{subfig:CBF}]{\includegraphics[width=\columnwidth]{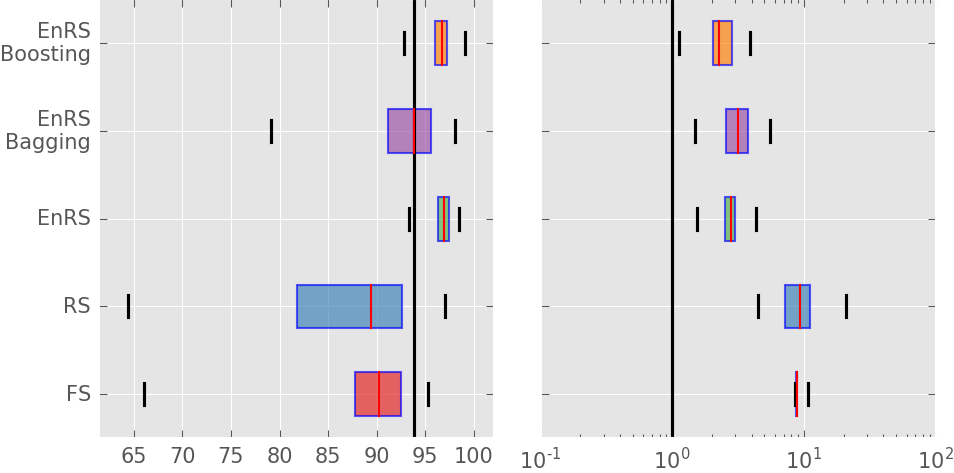}}
    \caption{Box plots showing achieved classification accuracy (left panel) and speed-up (right panel) for all evaluated datasets. The red lines show median values while minimum and maximum values are shown by black whiskers. The black line passing through the plot shows the values for YK-Shapelets algorithm.}
    \label{fig:tuned_all}
\end{figure*}

\begin{figure*}[p]
\ContinuedFloat
    \centering
     \subfloat[Coffee\label{subfig:Coffee}]{\includegraphics[width=\columnwidth]{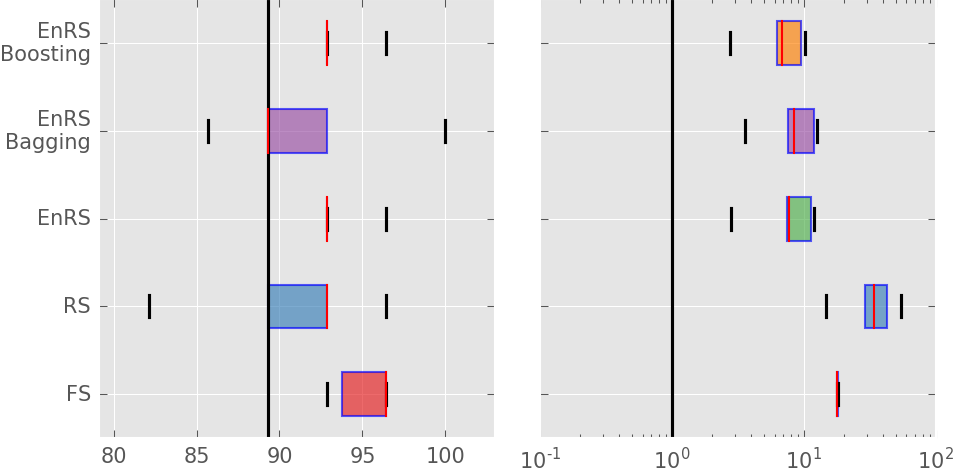}}
     \hspace{0.5cm}
     \subfloat[DiatomSizeReduction\label{subfig:DiatomSizeReduction}]{\includegraphics[width=\columnwidth]{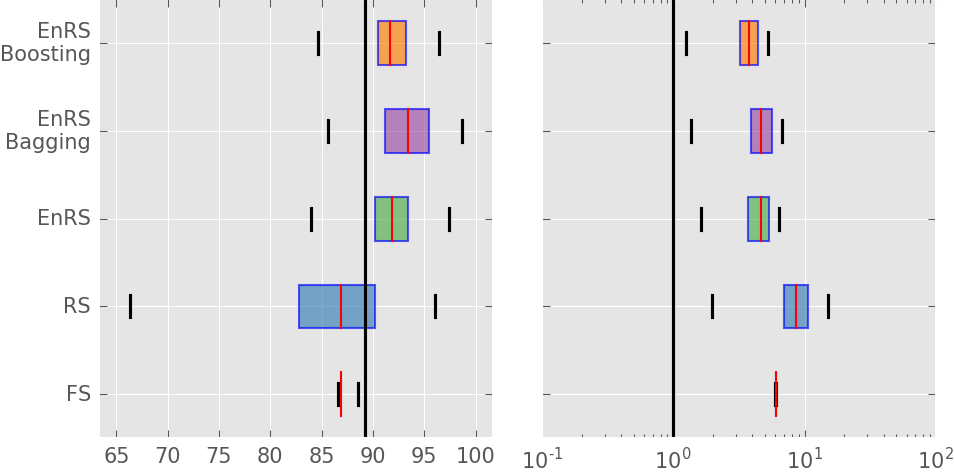}}

     \subfloat[DistalPhalanxOutlineAgeGroup\label{subfig:DistalPhalanxOutlineAgeGroup}]{\includegraphics[width=\columnwidth]{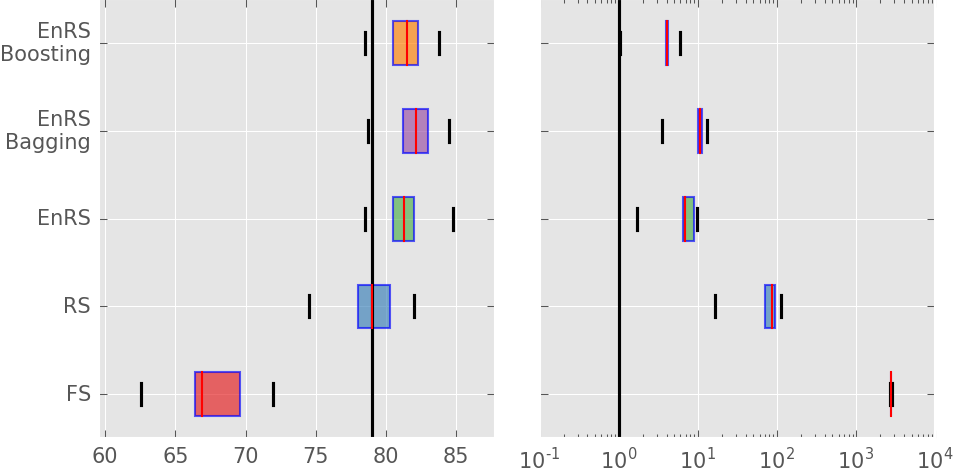}}
     \hspace{0.5cm}
     \subfloat[DistalPhalanxOutlineCorrect\label{subfig:DistalPhalanxOutlineCorrect}]{\includegraphics[width=\columnwidth]{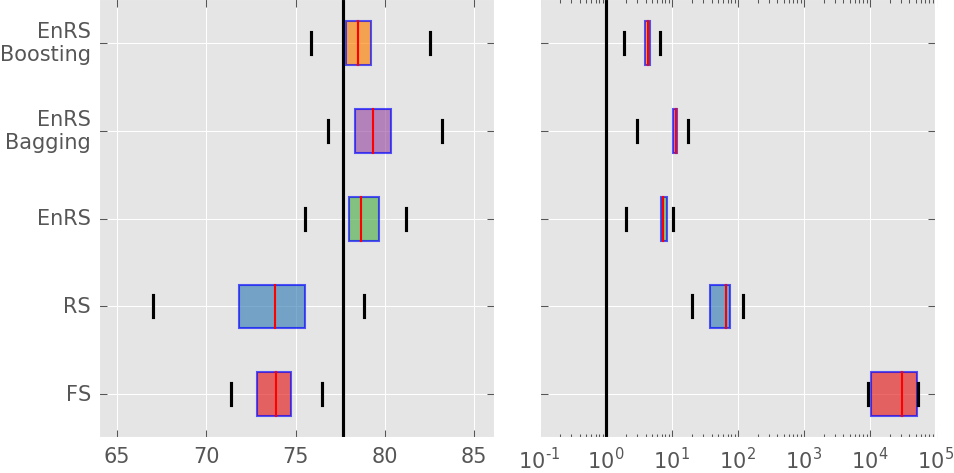}}

     \subfloat[DistalPhalanxTW\label{subfig:DistalPhalanxTW}]{\includegraphics[width=\columnwidth]{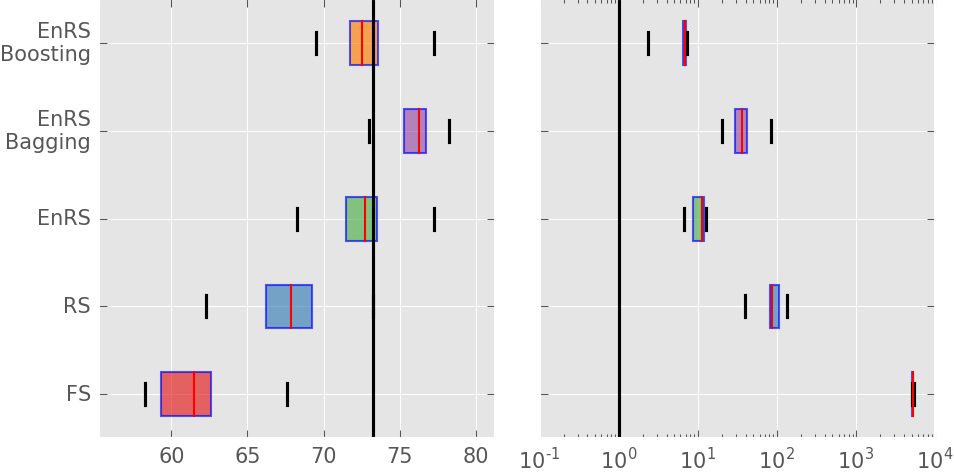}}
     \hspace{0.5cm}
     \subfloat[ECG200\label{subfig:ECG200}]{\includegraphics[width=\columnwidth]{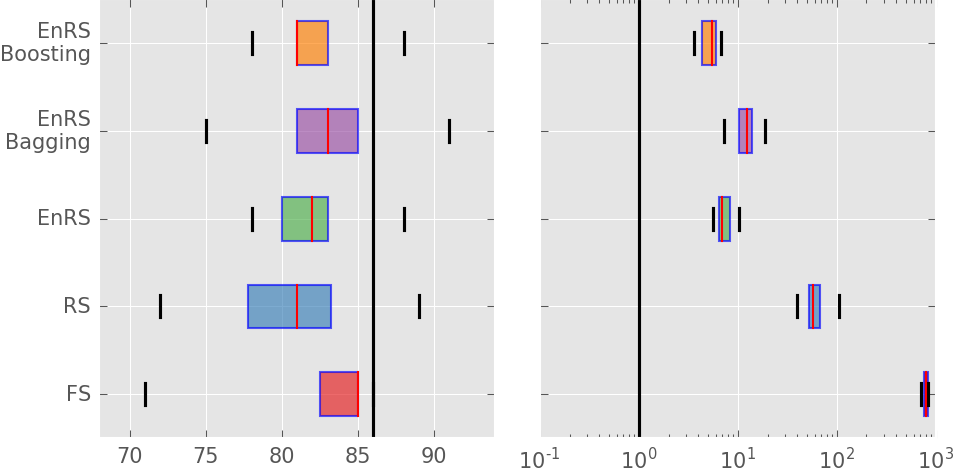}}

     \subfloat[ECGFiveDays\label{subfig:ECGFiveDays}]{\includegraphics[width=\columnwidth]{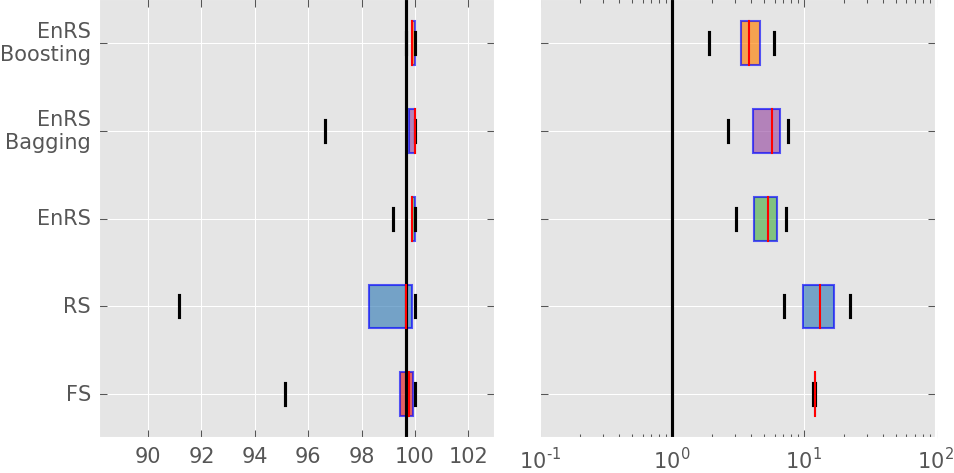}}
     \hspace{0.5cm}
     \subfloat[FaceAll\label{subfig:FaceAll}]{\includegraphics[width=\columnwidth]{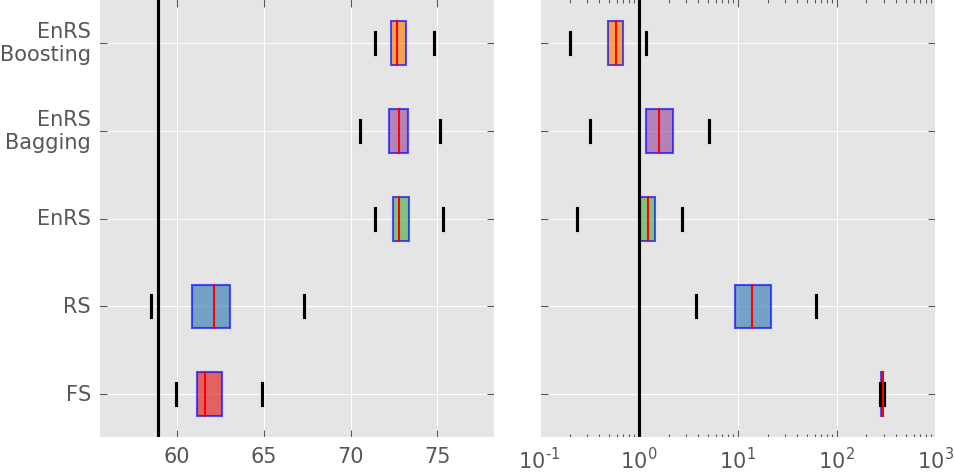}}
    \caption{Box plots showing achieved classification accuracy (left panel) and speed-up (right panel) for all evaluated datasets. The red lines show median values while minimum and maximum values are shown by black whiskers. The black line passing through the plot shows the values for YK-Shapelets algorithm.}
\end{figure*}

\begin{figure*}[p]
\ContinuedFloat
    \centering
     \subfloat[FaceFour\label{subfig:FaceFour}]{\includegraphics[width=\columnwidth]{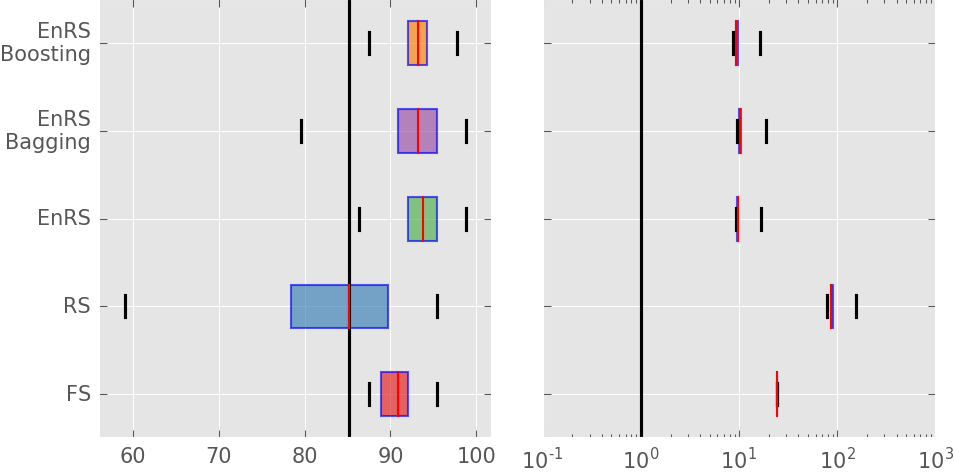}}
     \hspace{0.5cm}
     \subfloat[FacesUCR\label{subfig:FacesUCR}]{\includegraphics[width=\columnwidth]{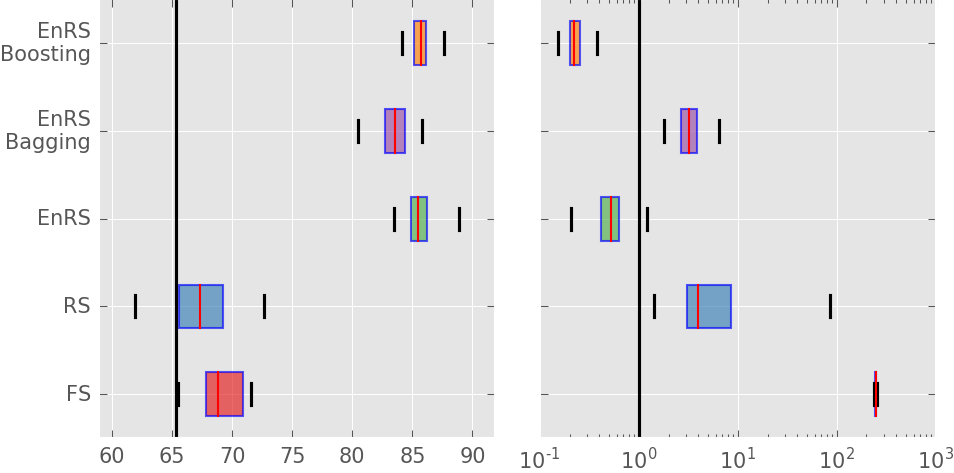}}

     \subfloat[FISH\label{subfig:FISH}]{\includegraphics[width=\columnwidth]{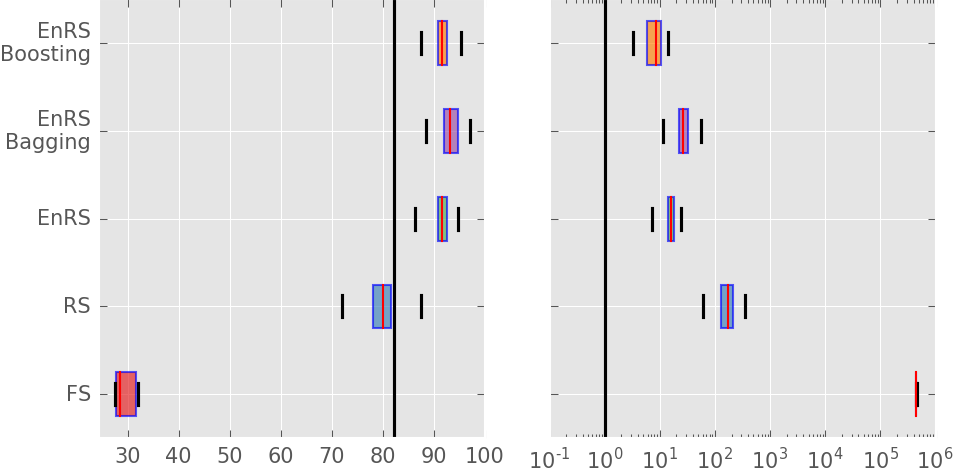}}
     \hspace{0.5cm}
     \subfloat[Gun\_Point\label{subfig:GunPoint}]{\includegraphics[width=\columnwidth]{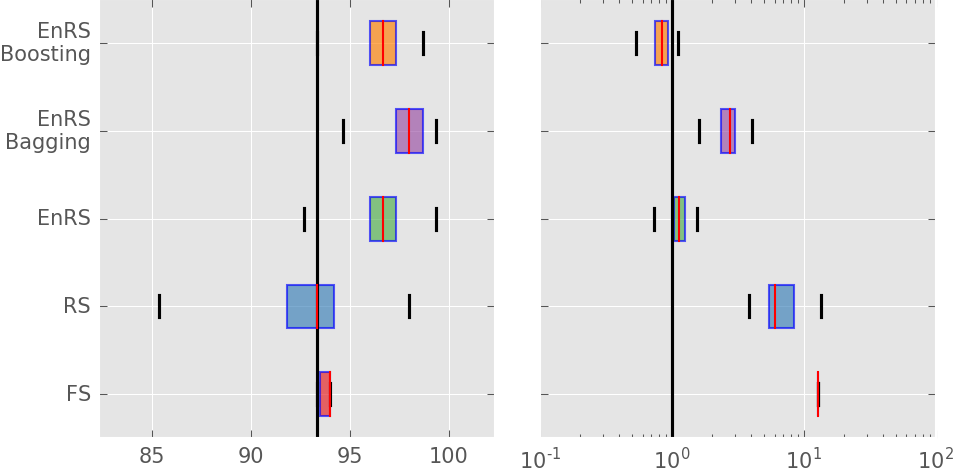}}

     \subfloat[Herring\label{subfig:Herring}]{\includegraphics[width=\columnwidth]{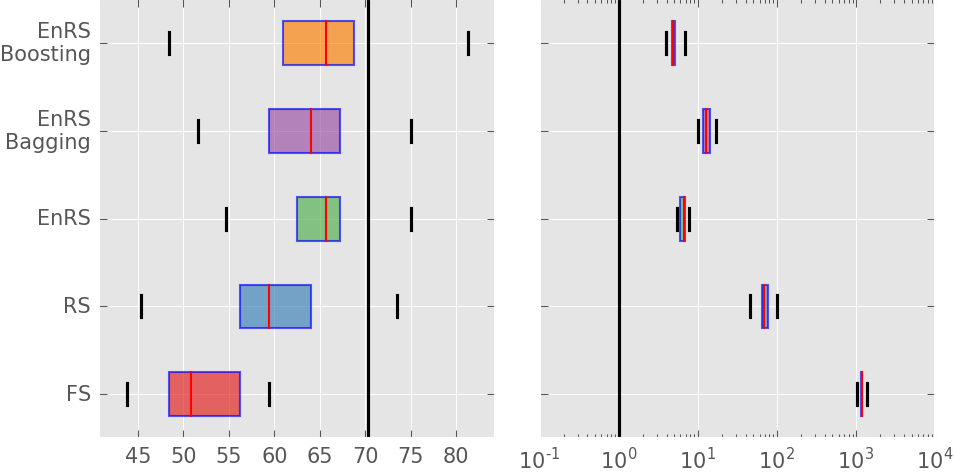}}
     \hspace{0.5cm}
     \subfloat[InsectWingbeatSound\label{subfig:InsectWingbeatSound}]{\includegraphics[width=\columnwidth]{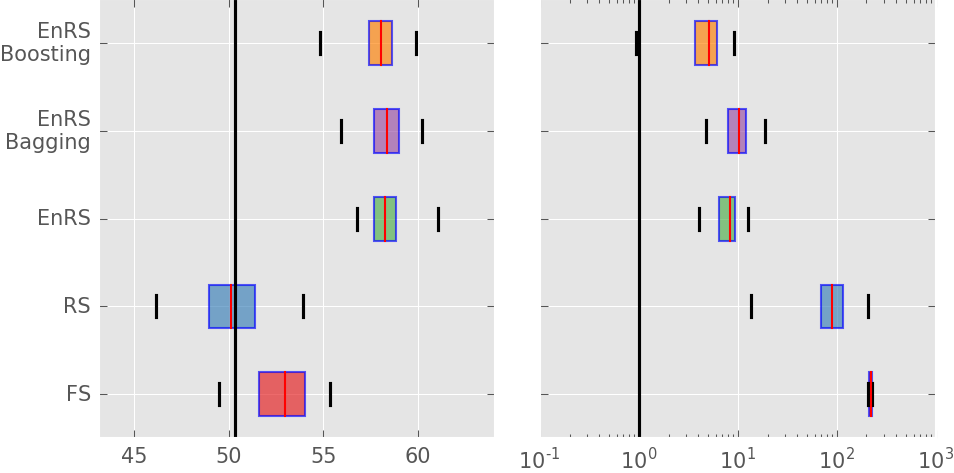}}

     \subfloat[ItalyPowerDemand\label{subfig:ItalyPowerDemand}]{\includegraphics[width=\columnwidth]{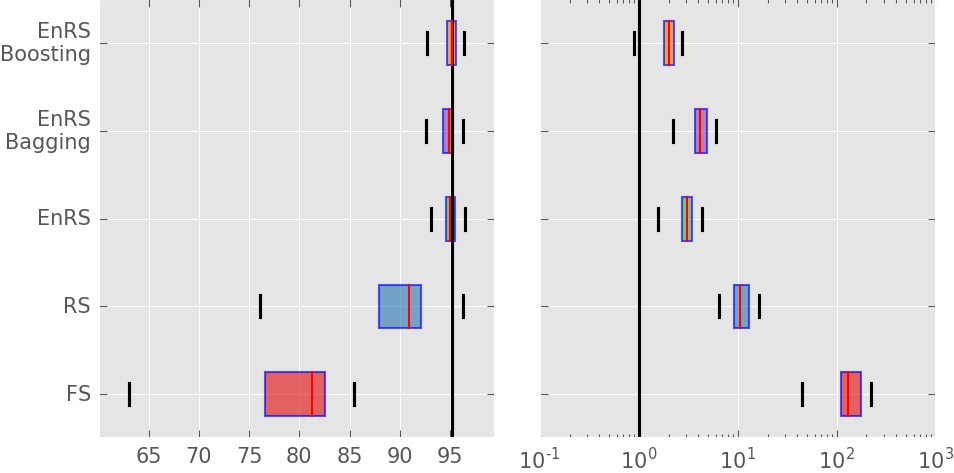}}
     \hspace{0.5cm}
     \subfloat[Lighting2\label{subfig:Lighting2}]{\includegraphics[width=\columnwidth]{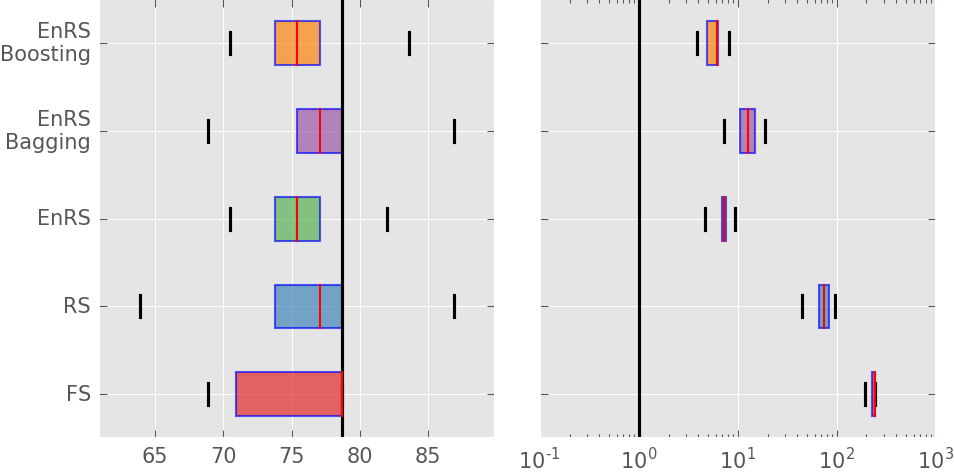}}
    \caption{Box plots showing achieved classification accuracy (left panel) and speed-up (right panel) for all evaluated datasets. The red lines show median values while minimum and maximum values are shown by black whiskers. The black line passing through the plot shows the values for YK-Shapelets algorithm.}
\end{figure*}

\begin{figure*}[p]
\ContinuedFloat
    \centering
     \subfloat[Lighting7\label{subfig:Lighting7}]{\includegraphics[width=\columnwidth]{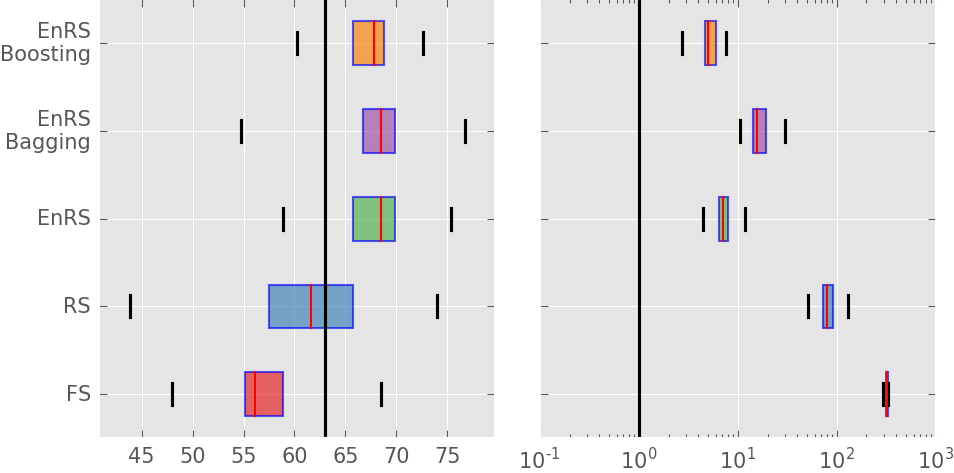}}
     \hspace{0.5cm}
     \subfloat[MALLAT\label{subfig:MALLAT}]{\includegraphics[width=\columnwidth]{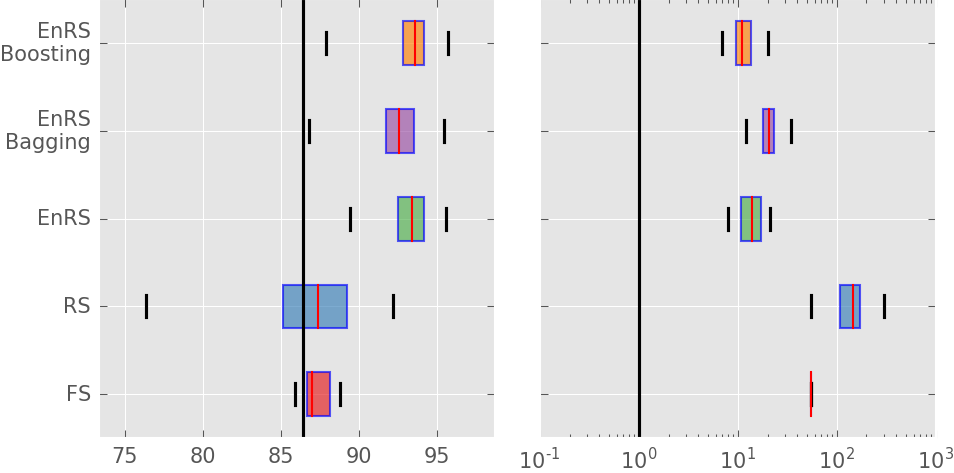}}

     \subfloat[Meat\label{subfig:Meat}]{\includegraphics[width=\columnwidth]{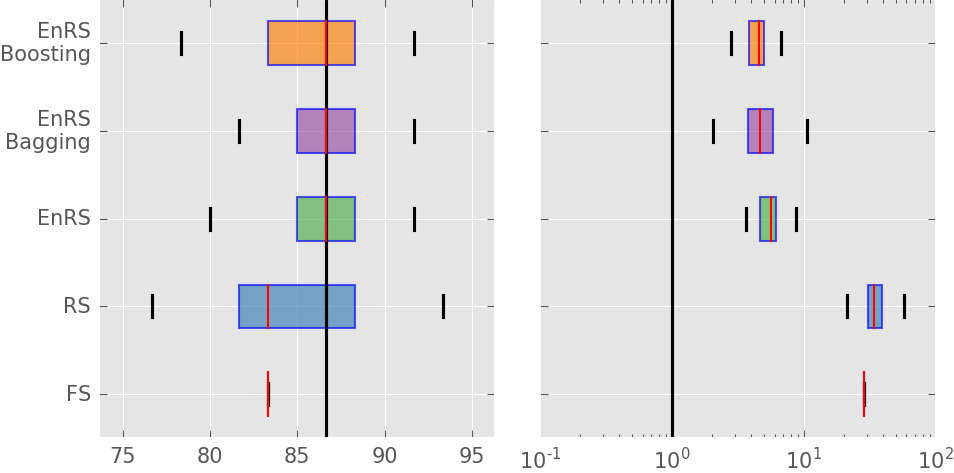}}
     \hspace{0.5cm}
     \subfloat[MiddlePhalanxOutlineAgeGroup\label{subfig:MiddlePhalanxOutlineAgeGroup}]{\includegraphics[width=\columnwidth]{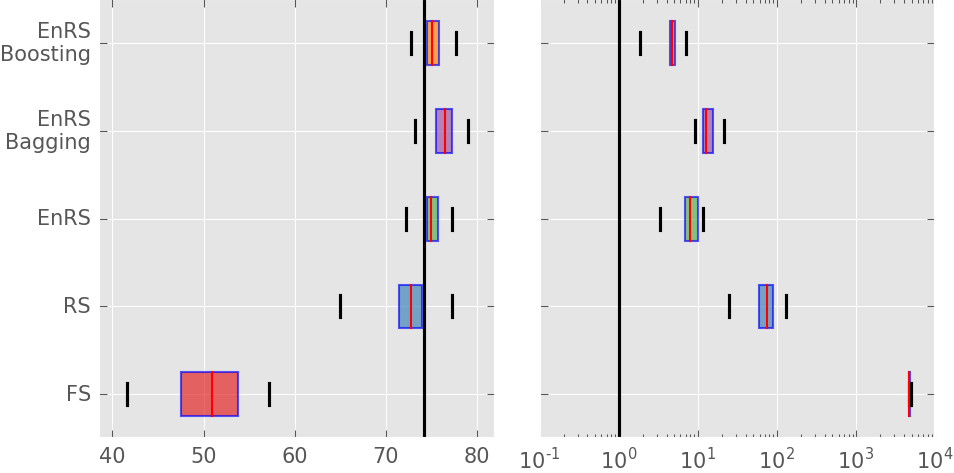}}

     \subfloat[MiddlePhalanxOutlineCorrect\label{subfig:MiddlePhalanxOutlineCorrect}]{\includegraphics[width=\columnwidth]{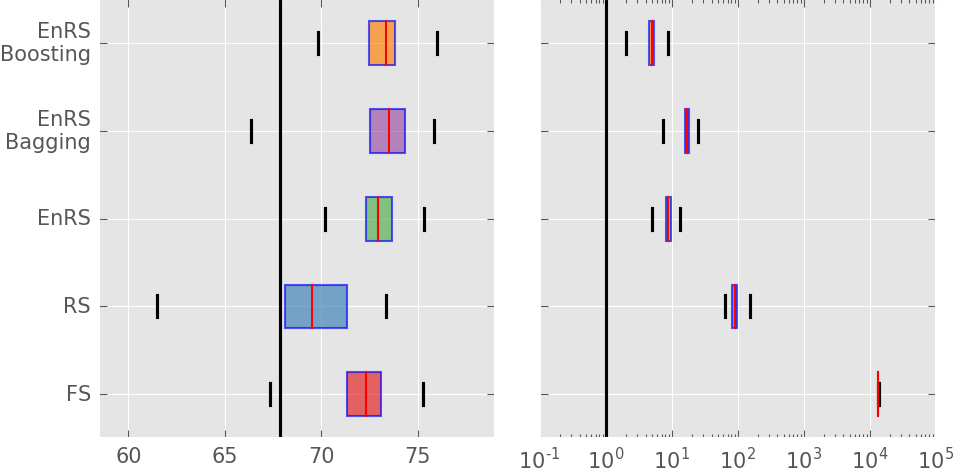}}
     \hspace{0.5cm}
     \subfloat[MiddlePhalanxTW\label{subfig:MiddlePhalanxTW}]{\includegraphics[width=\columnwidth]{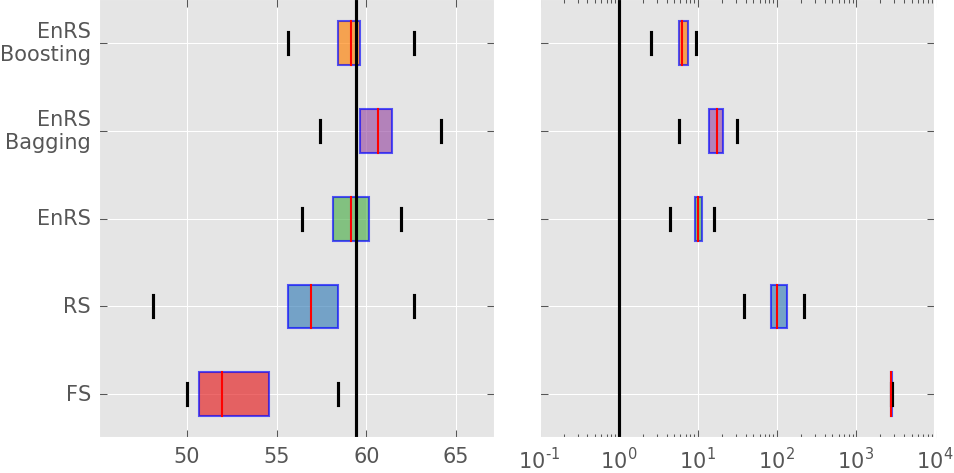}}

     \subfloat[MoteStrain\label{subfig:MoteStrain}]{\includegraphics[width=\columnwidth]{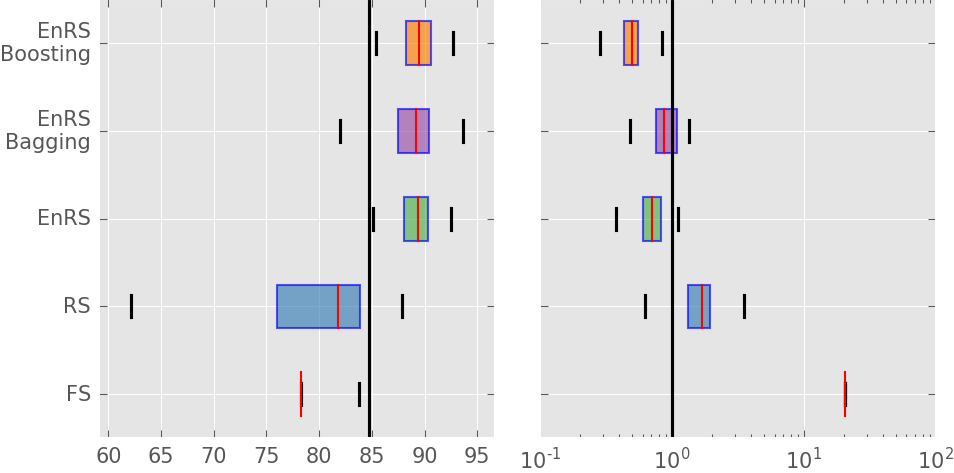}}
     \hspace{0.5cm}
     \subfloat[OliveOil\label{subfig:OliveOil}]{\includegraphics[width=\columnwidth]{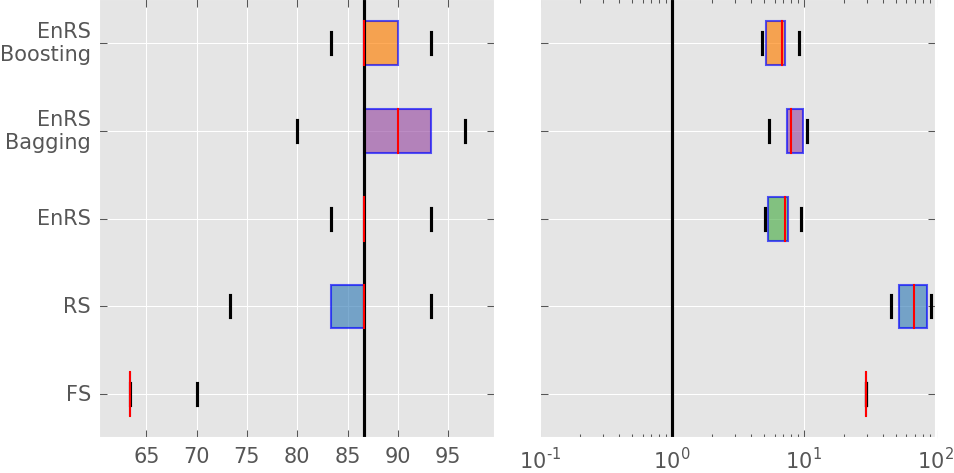}}
    \caption{Box plots showing achieved classification accuracy (left panel) and speed-up (right panel) for all evaluated datasets. The red lines show median values while minimum and maximum values are shown by black whiskers. The black line passing through the plot shows the values for YK-Shapelets algorithm.}
\end{figure*}

\begin{figure*}[p]
\ContinuedFloat
    \centering
     \subfloat[Plane\label{subfig:Plane}]{\includegraphics[width=\columnwidth]{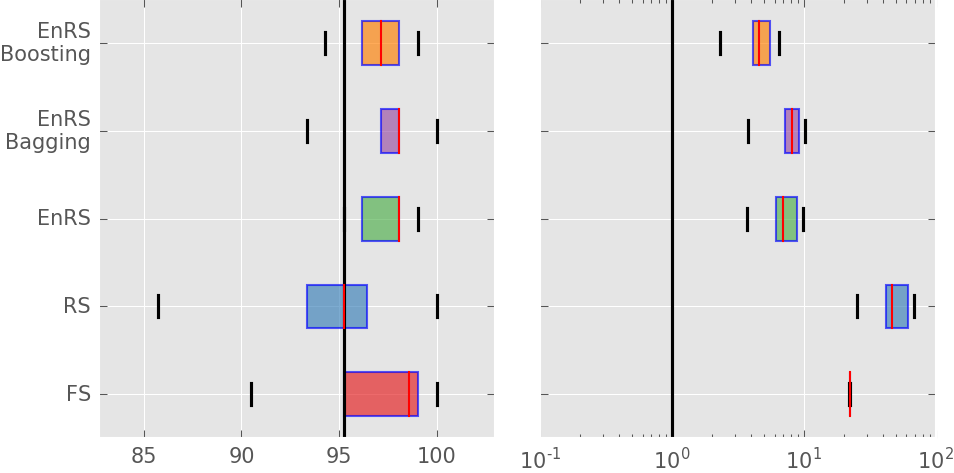}}
     \hspace{0.5cm}
     \subfloat[ProximalPhalanxTW\label{subfig:ProximalPhalanxTW}]{\includegraphics[width=\columnwidth]{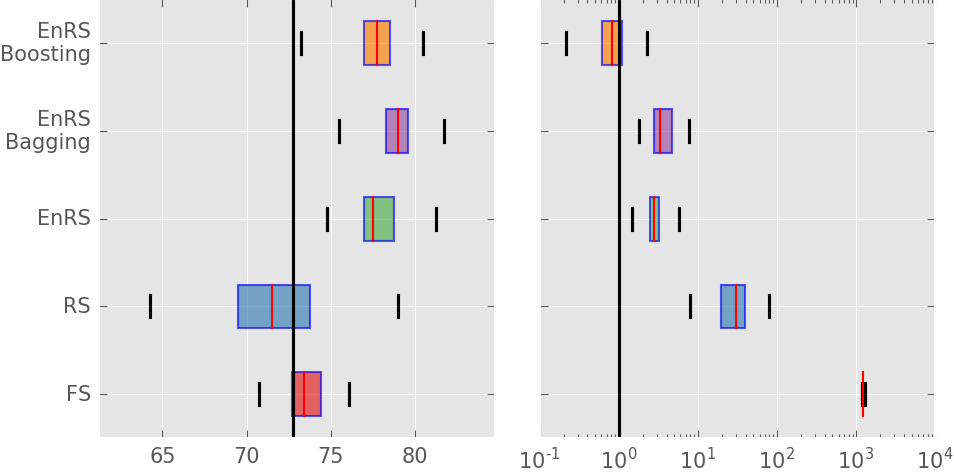}}

     \subfloat[ShapeletSim\label{subfig:ShapeletSim}]{\includegraphics[width=\columnwidth]{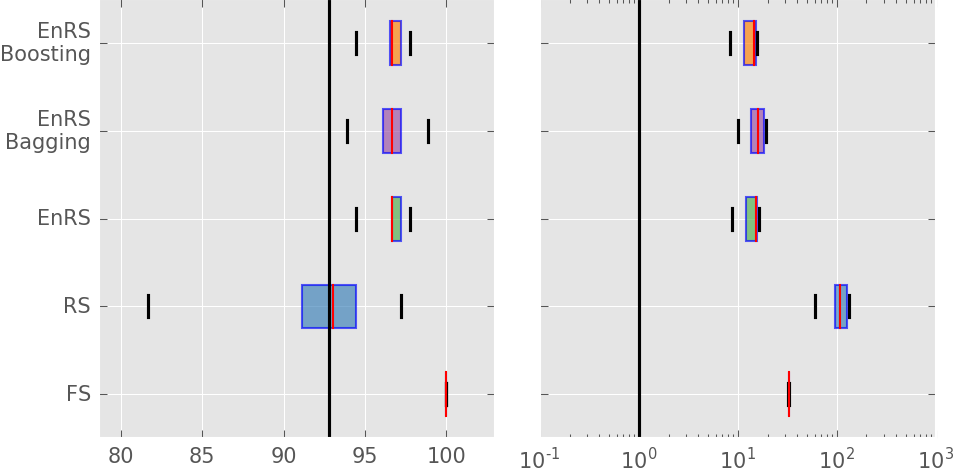}}
     \hspace{0.5cm}
     \subfloat[SonyAIBORobotSurface\label{subfig:SonyAIBORobotSurface}]{\includegraphics[width=\columnwidth]{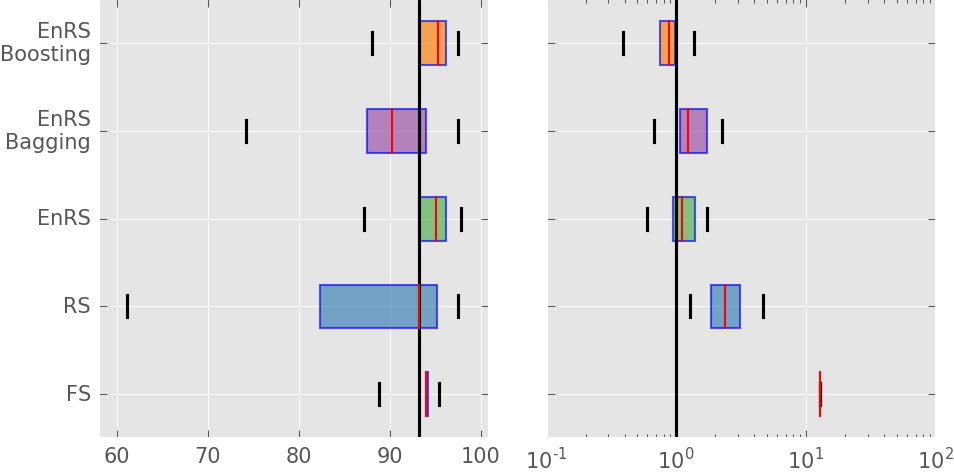}}

     \subfloat[SonyAIBORobotSurfaceII\label{subfig:SonyAIBORobotSurfaceII}]{\includegraphics[width=\columnwidth]{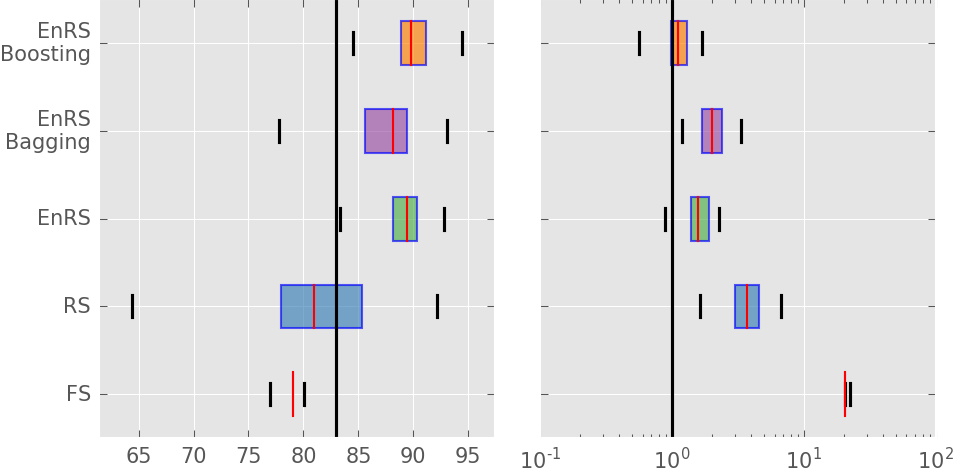}}
     \hspace{0.5cm}
     \subfloat[SwedishLeaf\label{subfig:SwedishLeaf}]{\includegraphics[width=\columnwidth]{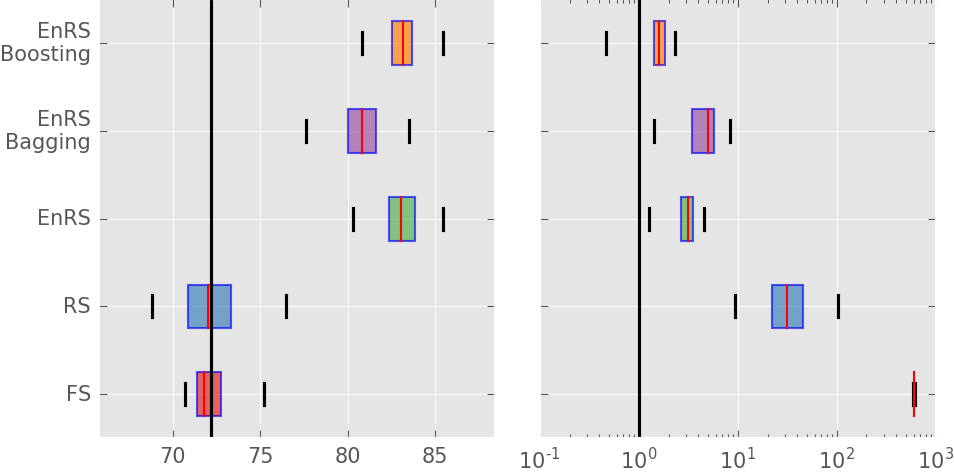}}

     \subfloat[Symbols\label{subfig:Symbols}]{\includegraphics[width=\columnwidth]{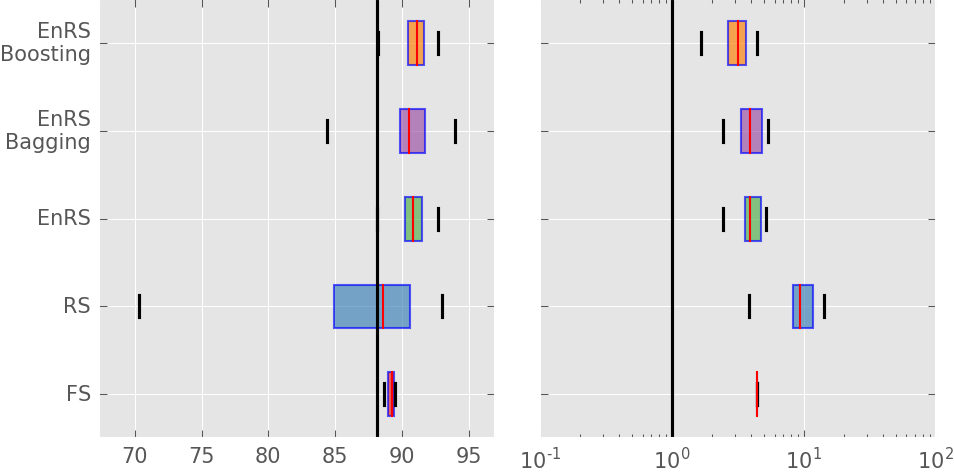}}
     \hspace{0.5cm}
     \subfloat[synthetic\_control\label{subfig:syntheticcontrol}]{\includegraphics[width=\columnwidth]{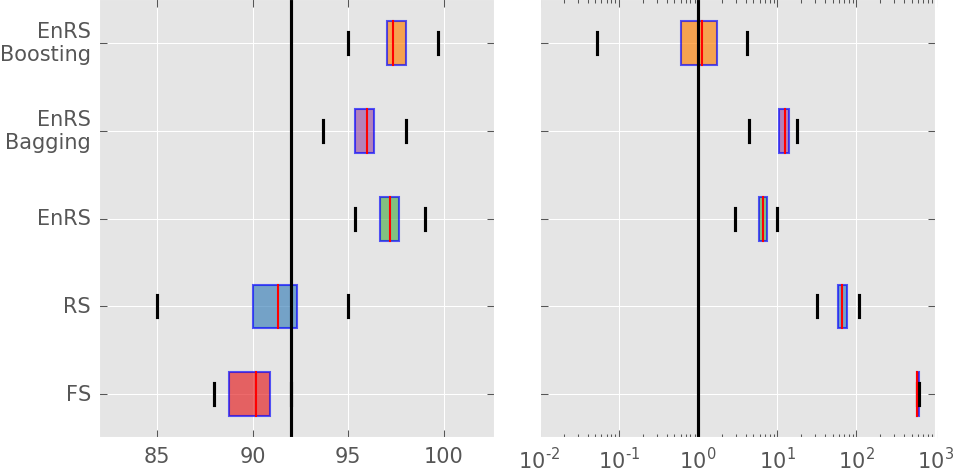}}
    \caption{Box plots showing achieved classification accuracy (left panel) and speed-up (right panel) for all evaluated datasets. The red lines show median values while minimum and maximum values are shown by black whiskers. The black line passing through the plot shows the values for YK-Shapelets algorithm.}
\end{figure*}

\begin{figure*}[p]
\ContinuedFloat
    \centering
     \subfloat[ToeSegmentation1\label{subfig:ToeSegmentation1}]{\includegraphics[width=\columnwidth]{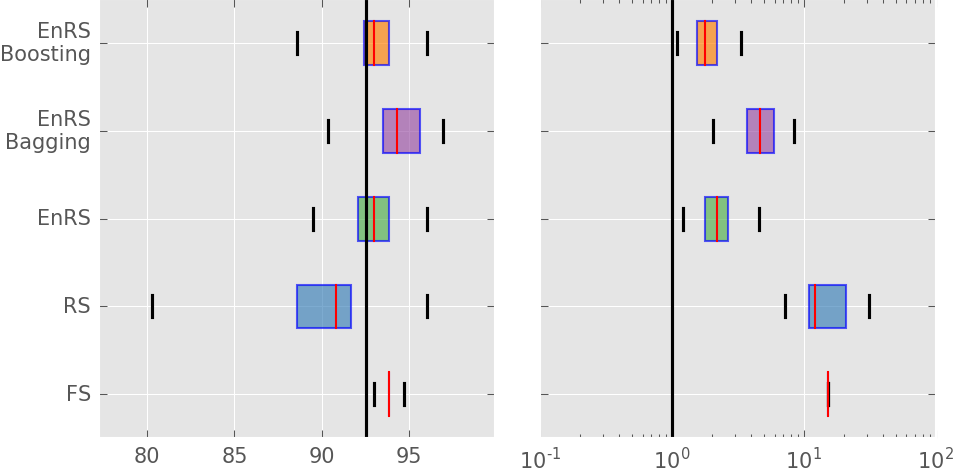}}
     \hspace{0.5cm}
     \subfloat[ToeSegmentation2\label{subfig:ToeSegmentation2}]{\includegraphics[width=\columnwidth]{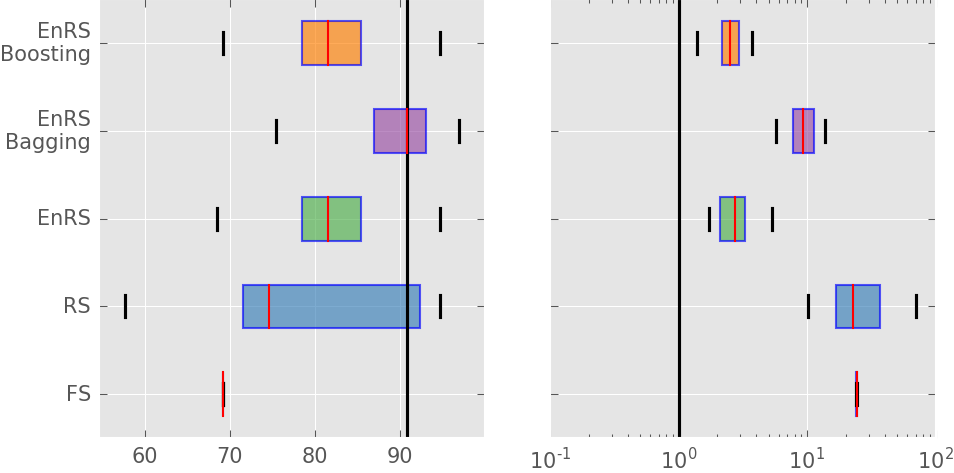}}

     \subfloat[Trace\label{subfig:Trace}]{\includegraphics[width=\columnwidth]{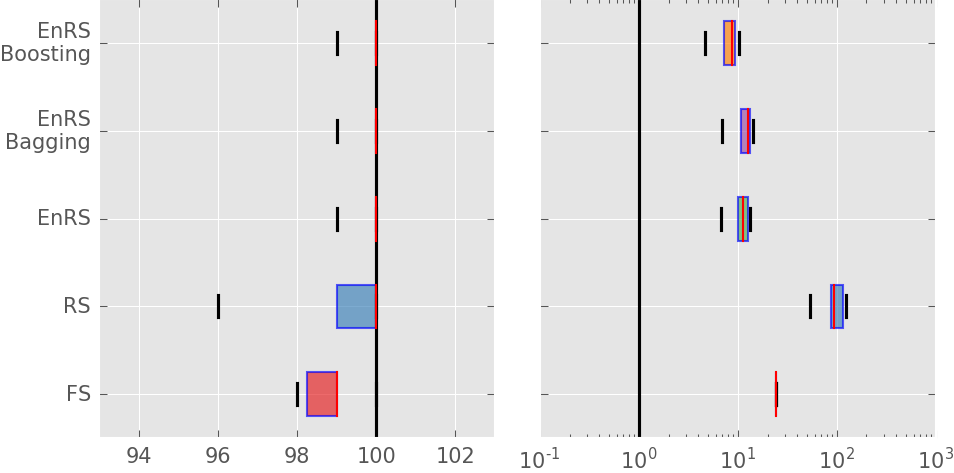}}
     \hspace{0.5cm}
     \subfloat[TwoLeadECG\label{subfig:TwoLeadECG}]{\includegraphics[width=\columnwidth]{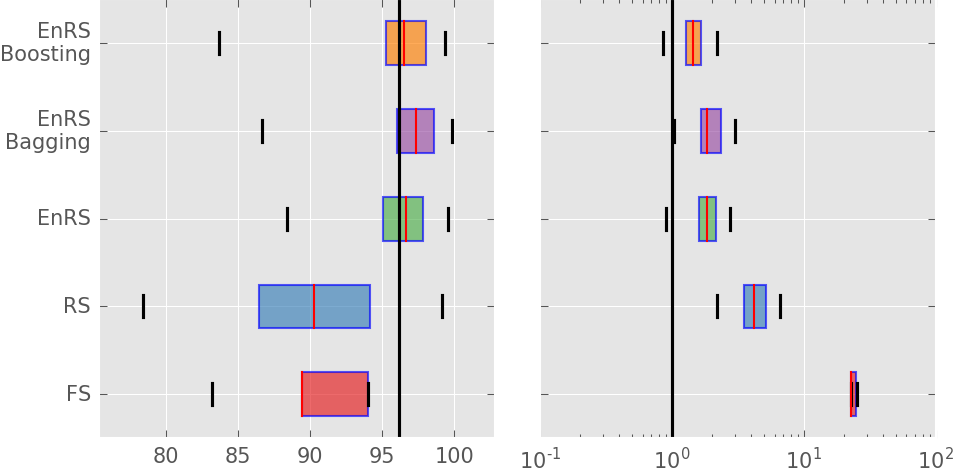}}

     \subfloat[Wine\label{subfig:Wine}]{\includegraphics[width=\columnwidth]{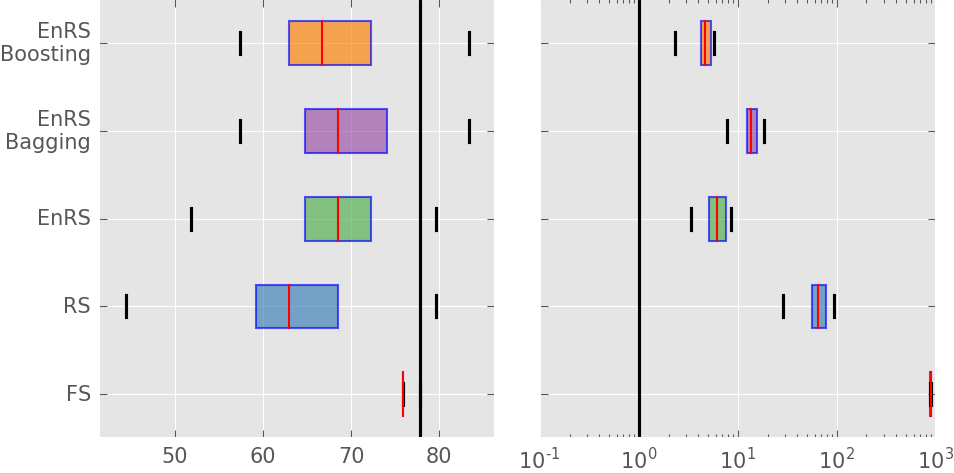}}
    \caption{Box plots showing achieved classification accuracy (left panel) and speed-up (right panel) for all evaluated datasets. The red lines show median values while minimum and maximum values are shown by black whiskers. The black line passing through the plot shows the values for YK-Shapelets algorithm.}
\end{figure*}

\subsection{EnRS-Bagging vs.\ gRSF}\label{subsec:EnRSBagging_vs_gRSF}
The Generalized Random Shapelet Forests or gRSF algorithm is very similar to the approach we call Ensembles of Random-Shapelets using Bagging or EnRS-Bagging, so we compared the classification accuracy of the two algorithms.
We used the Java implementation provided by the authors of gRSF and evaluated all the datasets evaluated in our other experiments.
The $minLen$ and $maxLen$ parameters were set to the best parameters reported in the gRSF paper.
The results reported in the gRSF paper were obtained by setting the ensemble size to 500, while we performed all our experiments with merely 10 models per ensemble, so we also performed the experiments for gRSF using 10 models per ensemble to make a fair comparison between the two algorithms.
Figure \ref{fig:gRSFvsEnRSBagging} shows the critical differences diagram and the average ranks for the classification accuracy results for the two algorithms.
The bagging ensemble has a slightly better average rank than gRSF and the Nemenyi test does not find them significantly different at a $p=0.05$ significance level.
This slight difference can be explained due to the smaller number of candidates evaluated by the gRSF algorithm.
By default, the number of candidates sampled by the gRSF algorithm at each node is equal to $\sqrt{\frac{1}{2}m(m+1)}$, where $m$ is the length of the time series.
This number turns out to be even smaller than 1\% of the total shapelet candidates used in our experiments.
In fact, for at least the root nodes, using this number of candidates will always be smaller than 1\% of the total shapelet candidates for all the datasets in the UCR Time Series Archive.
This observation points to the fact that even a lesser number of candidates can yield very promising results for the shapelet-based ensembles.
Using an even smaller percentage of candidates will also lead to better run times.

\begin{figure}[tb]
    \centering
    \includegraphics[width=\columnwidth]{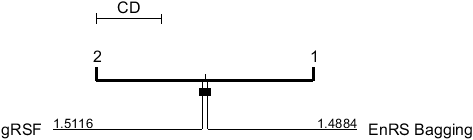}
    \caption{Average ranks for Ensembles of Random-Shapelets with Bagging (EnRS-Bagging) and Generalized Random Shapelet Forests (gRSF) using ten models per ensemble and 1\% sampling for Random-Shapelets models and the \textit{default} number of sampled candidates for gRSF. The Nemenyi test is performed at the $p=0.05$ significance level.}
    \label{fig:gRSFvsEnRSBagging}
\end{figure}

\section{Discussion}\label{section:discussion}
The EnRS-Bagging approach is the fastest out of the three ensemble approaches tested in our experimentation while EnRS-Boosting is the slowest.
The EnRS-Bagging approach also has a higher number of overall wins head-to-head with the other two ensemble approaches.
EnRS-Bagging performs better in classification accuracy and runtime because of the duplication of instances in the dataset used for training.
The duplicated instances allow the candidate pruning strategy to quickly identify good or bad candidates.
Therefore, bagging allows the algorithm to run faster.
Using duplicated instances in the training dataset also introduces a bias towards the instances of the majority class.
This leads to an early extraction of the shapelet specific to the instances of the majority class and allows the algorithm to effectively split the dataset early and then search for shapelets for the other instances.
This also makes the process efficient.
The EnRS-Boosting approach performs slower because the weighting of instances increases the computation required for performing data splits and hence the candidate pruning.
Since the candidate pruning strategy creates optimistic splits in each call, this becomes a limiting factor for the EnRS-Boosting approach.

The Fast-Shapelets algorithm is also a heuristic method and can be a possible candidate for the base learner in the ensemble learning approach.
We experimented with this approach as well, however, the Fast-Shapelets algorithm does not provide much diversity in the models, which makes its use in ensembles less effective than the Random-Shapelets algorithm.

\section{Conclusion}\label{section:conclusions}
We proposed an ensemble learning approach using Random-Shapelets algorithm as a base classifier for shapelets based classification.
The use of an inexpensive but reasonably accurate base learner proves to be highly beneficial.
The benefits of the proposed method are twofold and include better classification accuracy and reduced computational effort.
Better classification accuracy was achieved for almost all the evaluated datasets, while the run time was reduced in all cases.
The simplicity and added benefits of the approach make it very suitable for shapelet discovery and classification. 
Using bagging can reduce the required computation, however, in some cases the classification accuracy of the obtained model is slightly worse than the ensemble of Random-Shapelets classifiers trained on the original training dataset, albeit not significantly.


Currently, the Random-Shapelets algorithm can only evaluate candidates with a sampling ratio set at start of the process.
The possibility of changing the fraction of evaluated candidates and use the results in an additive fashion to the already obtained results could prove beneficial.
This would require some book keeping about the already evaluated candidates and the obtained results, but if the storage requirements can be kept low, this could prove as a refinement step to an approximate solution.
Another future research avenue could be the use of Random-Shapelets based classification models trained using randomly chosen window length parameters and combining the models in an ensemble.
This should, theoretically, enhance the diversity of the individual models and also remove the need to perform parameter tuning before model generation.

\bibliographystyle{abbrv}
\bibliography{bibfile}

\begin{thebibliography}{10}

\bibitem{Bagnall2014NNEval}
A.~Bagnall and J.~Lines.
\newblock {An Experimental Evaluation of Nearest Neighbour Time Series
  Classification}.
\newblock {\em CoRR}, 1406.4757, 2014.

\bibitem{Breiman1996bagging}
L.~Breiman.
\newblock Bagging predictors.
\newblock {\em Machine learning}, 24(2):123--140, 1996.

\bibitem{Ding2008CompDistMeasure}
H.~Ding, G.~Trajcevski, P.~Scheuermann, X.~Wang, and E.~Keogh.
\newblock {Querying and mining of time series data: experimental comparison of
  representations and distance measures}.
\newblock {\em Proceedings of the VLDB Endowment}, 1(2):1542--1552, 2008.

\bibitem{freund1995desicion}
Y.~Freund and R.~E. Schapire.
\newblock A desicion-theoretic generalization of on-line learning and an
  application to boosting.
\newblock In {\em Computational learning theory}, pages 23--37. Springer, 1995.

\bibitem{Hansen1990NNEnsembles}
L.~Hansen and P.~Salamon.
\newblock Neural network ensembles.
\newblock {\em IEEE Transactions on Pattern Analysis and Machine Intelligence},
  12(10):993--1001, 1990.

\bibitem{Karlsson2016gRSF}
I.~Karlsson, P.~Papapetrou, and H.~Bostr{\"{o}}m.
\newblock {Generalized random shapelet forests}.
\newblock {\em Data Mining and Knowledge Discovery}, 30(5):1053--1085, 2016.

\bibitem{Lin2007SAX}
J.~Lin, E.~Keogh, L.~Wei, and S.~Lonardi.
\newblock {Experiencing SAX: a novel symbolic representation of time series}.
\newblock {\em Data Mining and Knowledge Discovery}, 15(2):107--144, 2007.

\bibitem{Lines2012ShapeletTransform}
J.~Lines, L.~M. Davis, J.~Hills, and A.~Bagnall.
\newblock {A shapelet transform for time series classification}.
\newblock In {\em Proceedings of the 18th ACM SIGKDD international conference
  on Knowledge discovery and data mining - KDD '12}, pages 289--297, New York,
  New York, USA, 2012. ACM Press.

\bibitem{Mueen2011Logical}
A.~Mueen, E.~Keogh, and N.~Young.
\newblock {Logical-Shapelets: An Expressive Primitive for Time Series
  Classification}.
\newblock In {\em Proceedings of the 17th ACM SIGKDD international conference
  on Knowledge discovery and data mining - KDD '11}, page 1154, New York, New
  York, USA, 2011. ACM Press.

\bibitem{Rakthanmanon2013FastShapelets}
T.~Rakthanmanon and E.~Keogh.
\newblock {Fast Shapelets: A Scalable Algorithm for Discovering Time Series
  Shapelets}.
\newblock In {\em Proceedings of the 2013 SIAM International Conference on Data
  Mining}, pages 668--676, Philadelphia, PA, may 2013. Society for Industrial
  and Applied Mathematics.

\bibitem{Renard2015RandomShapelets}
X.~Renard, M.~Rifqi, W.~Erray, and M.~Detyniecki.
\newblock {Random-shapelet: An algorithm for fast shapelet discovery}.
\newblock In {\em 2015 IEEE International Conference on Data Science and
  Advanced Analytics (DSAA)}, pages 1--10. IEEE, Oct 2015.

\bibitem{Sakurai2005BRAID}
Y.~Sakurai, S.~Papadimitriou, and C.~Faloutsos.
\newblock Braid: Stream mining through group lag correlations.
\newblock In {\em Proceedings of the 2005 ACM SIGMOD international conference
  on Management of data}, pages 599--610. ACM, 2005.

\bibitem{Ye2009Shapelets}
L.~Ye and E.~Keogh.
\newblock {Time Series Shapelets: A New Primitive for Data Mining}.
\newblock In {\em Proceedings of the 15th ACM SIGKDD international conference
  on Knowledge discovery and data mining - KDD '09}, pages 947--956, New York,
  New York, USA, 2009. ACM Press.

\end{thebibliography}
%
%
\end{document}